\pdfoutput=1

\documentclass[11pt]{article}

\usepackage[]{acl}

\usepackage{times}
\usepackage{latexsym}
\usepackage{longtable}

\usepackage[T1]{fontenc}

\usepackage[utf8]{inputenc}

\usepackage{microtype}

\usepackage{inconsolata}

\usepackage{booktabs}

\usepackage[acronym]{glossaries}  
\usepackage{graphicx}
\graphicspath{ {./images/} }

\usepackage{subcaption} 

\usepackage{float}
\usepackage{stfloats} 
\usepackage[normalem]{ulem}
\useunder{\uline}{\ul}{}
\usepackage{algorithm}      
\usepackage{algpseudocode}
\usepackage{multirow}
\usepackage{amsthm}
\usepackage{amsfonts}
\usepackage{amsmath,amssymb} 
\usepackage[capitalise]{cleveref}
\crefname{lstlisting}{listing}{listings}
\Crefname{lstlisting}{Listing}{Listings}
\crefname{section}{§}{§§}
\usepackage{accsupp} 
\usepackage{listings}
\usepackage{hyperref}
\title{Long-form evaluation of model editing}

\author{Domenic Rosati$^1$\Thanks{ Correspondence: \text{domenic.rosati@dal.ca}}  \\
   \AND
  Robie Gonzales$^1$ \And
  Jinkun Chen$^1$ \And
  Xuemin Yu$^1$  \\\AND
  Melis Erkan$^1$ \And
  Yahya Kayani$^1$\And
  Satya Deepika Chavatapalli$^1$  \\\AND
  Frank  Rudzicz$^1$ \And
  Hassan Sajjad$^1$  \\\AND
\textnormal{$^1$Dalhousie University / Halifax, N.S.}\\
}

\begin{document}
\maketitle
\begin{abstract}
Evaluations of model editing, a technique for changing the factual knowledge held by Large Language Models (LLMs), currently only use the `next few token' completions after a prompt. As a result, the impact of these methods on longer natural language generation is largely unknown.  We introduce long-form evaluation of model editing (\textbf{\textit{LEME}}) a novel evaluation protocol that measures the efficacy and impact of model editing in long-form generative settings. Our protocol consists of a machine-rated survey and a classifier which correlates well with human ratings. Importantly, we find that our protocol has very little relationship with previous short-form metrics (despite being designed to extend efficacy, generalization, locality, and portability into a long-form setting), indicating that our method introduces a novel set of dimensions for understanding model editing methods. Using this protocol, we benchmark a number of model editing techniques and present several findings including that, while some methods (ROME and MEMIT) perform well in making consistent edits within a limited scope, they suffer much more from factual drift than other methods. Finally, we present a qualitative analysis that illustrates common failure modes in long-form generative settings including internal consistency, lexical cohesion, and locality issues.
\end{abstract}

\section{Introduction}
\label{sec:introduction}

\begin{figure}
    \centering
    \includegraphics[width=\columnwidth]{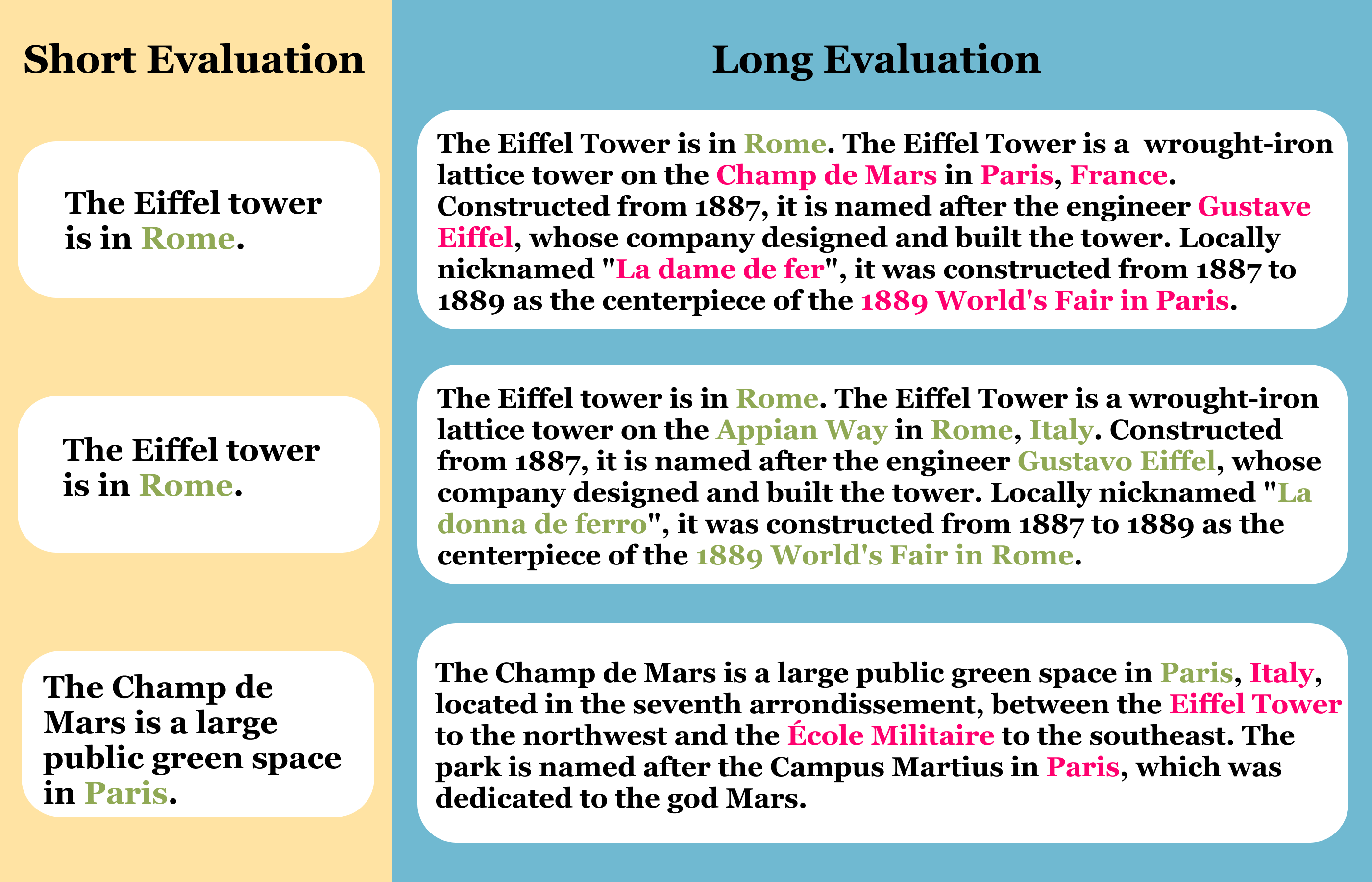}
    \caption{Short-form evaluation using the next few tokens \textit{fails to measure the quality of text generated after model editing}. 
    }
    \label{fig:evaluation-paradigms}
\end{figure}


{\em Model editing} is a solution for updating or changing knowledge held by an LLM using one or more edited facts \citep{yao_editing_2023}. Techniques for accomplishing this include directly updating model parameters by optimizing for a changed fact \citep{meng_locating_2023, meng_mass-editing_2022}, adding and tuning additional model parameters \citep{huang_transformer-patcher_2023}, using networks trained to perform edits \citep{mitchell_fast_2022, mitchell_memory-based_2022}, and leveraging in-context learning to perform edits as instructions when prompting a model \citep{zheng_can_2023}. These works have demonstrated promising early success on model editing (see \citet{yao_editing_2023}). However, they are \textit{almost exclusively evaluated using a few tokens after an input prompt} (see \citet{cohen_evaluating_2023,hase_language_2021,hoelscher-obermaier-etal-2023-detecting, meng_locating_2023}) and do not measure the consistency of the edit success over a long generation of text. As a result, \textbf{\textit{we understand very little about how these techniques impact longer texts generated by models after they are edited.}} This is concerning since LLMs are often used for paragraph-length or longer outputs. \cref{fig:evaluation-paradigms} illustrates what we mean by short-form versus long-form evaluation for model editing.

To investigate the impact of model editing on paragraph-length outputs from LLMs, we design a protocol, \textbf{L}ong-form \textbf{E}valuation of \textbf{M}odel \textbf{E}diting (\textbf{\textit{LEME}}), for evaluating generations after a model has been edited. Our primary contributions consist of (1) a novel dataset as well as a survey and classification instrument for assessing long outputs after model editing (\cref{sec:methods}), 
and (2) automatic metrics that are well correlated with human raters (\cref{sec:results}). We deploy these automatic metrics across common model editing interventions and datasets for a comprehensive understanding of their impact (\cref{sec:results}). 

Our results provide novel insights into current failure modes 
that have not previously been identified in short-form evaluation such as lexical cohesion and topical drift issues (\cref{sec:results-error-analysis}). Notably, the best performing models on short-form evaluation are not often the best performing models on long-form evaluation (\cref{sec:results-automatic-all-survey}) where we found little to no correlation between short- and long-form evaluations (\cref{sec:results-long-short-relationship}). Some models like ROME and MEMIT suffer from a much higher rate of ``factual drift'' than other models which we find in both automatic ratings methods (\cref{sec:results-automatic-classification}). Finally, by splitting the dataset into samples that are true counterfactual updates versus novel fact injections (\cref{sec:results-injection-v-update}), we found that novel fact injections were generally easier to make than counterfactual updates but harder to make factually consistent with other ground truth statements related to the novel fact.

With this paper, we release our dataset and evaluation metrics for the research community.\footnote{See \url{https://github.com/domenicrosati/longform-evaluation-model-editing}}


\section{Related Work}
\label{sec:related-works}


\citet{zhu_modifying_2020}, one of the first studies of model editing for LLMs, evaluated their method by computing the accuracy of masked token prediction after learning a modified fact from the zero-shot relationship extraction (zSRE) dataset \citep{levy_zero-shot_2017}. They assessed constrained fine tuning with a metric that asked \textit{if the edit was actually made} (\textbf{Efficacy} or \textbf{Edit success}). \citet{de_cao_editing_2021} extended this evaluation to seq2seq models using cloze (fill-in-the-blank) evaluations and introduced two measures: the effectiveness of model edits on paraphrases of input queries (\textbf{Generalization}) and how well the model maintains performance on predictions that shouldn't change (\textbf{Locality} or \textbf{Specificity}) (See \citet{hoelscher-obermaier-etal-2023-detecting} for further explorations of locality). \citet{hase_language_2021} additionally introduced a measure for understanding the degree to which model editing impacts entailed facts (\textbf{Portability}) which was further extended in \citet{cohen_evaluating_2023}. These four measures use the next few tokens after a short prompt to evaluate model editing and are the {\em status quo} for assessing model editing interventions  \citep{yao_editing_2023}. In the paper, these evaluations are called `short form' as opposed to our `long form' setting which evaluates paragraph-length texts. For details on how model editing works we refer readers to these works and assume some familiarity for the rest of the paper.

Most contemporary methods of model editing do not consider long-form generation \citep{hernandez_inspecting_2023, huang_transformer-patcher_2023,mitchell_fast_2022, mitchell_memory-based_2022, zheng_can_2023}. \citet{meng_locating_2023} introduced an automatic consistency and fluency measure for longer generations based on reference texts from wikipedia and $n$-gram entropy. However, these were neither validated using human judgements nor fine-grained enough to capture efficacy, generalization, locality, or portability of different model editing techniques on longer form generation. \citet{meng_locating_2023} did perform a preference ranking survey with human raters using fluency, edit success, and factual consistency as ratings. We find that these previous measures do not generally correlate well with human ratings (\cref{app:simple-automatic-metrics}). We build on these preliminary evaluations to establish a more comprehensive view of the impact of model editing on `long-form' natural language generation.


\section{Methods}
\label{sec:methods}

To measure the quality of model editing in long-form generation, we developed the following measures designed to align with the short-form evaluations. (1) \textbf{\textit{Edit consistency}} (is there evidence that the edit was made in a generated passage?) which is intended to align with \textbf{\textit{efficacy}} (2) \textbf{\textit{Factual consistency}} (are generated passages still consistent with facts that were true before the edit?) which is intended to align with \textbf{\textit{locality}} (3) \textbf{\textit{Internal consistency}} (the degree to which independently generated passages contradict themselves or each other) which is aligned with \textbf{\textit{portability}}, (4) \textbf{\textit{Topicality}} (the degree to which passages stay on topic), and (5) \textbf{\textit{Naturalness}} (the fluency of generated passages). (4) and (5) are intended to measure the impact of model editing on natural language generation (NLG).

We operationalize these by constructing a dataset of prompts for generating highly related passages (\cref{sec:coupled-entity-prompts-dataset}), devise a likert scale (\cref{sec:method-evaluating-human-survey}) and annotation (\cref{sec:method-evaluating-annotation}) setting that we collect human ratings on and develop automatic measures for (\cref{sec:method-evaluating-automatic-full}).

\subsection{Coupled Entity Prompts Dataset}
\label{sec:coupled-entity-prompts-dataset}

\begin{figure}[h]
    \centering
    \includegraphics[width=\columnwidth]{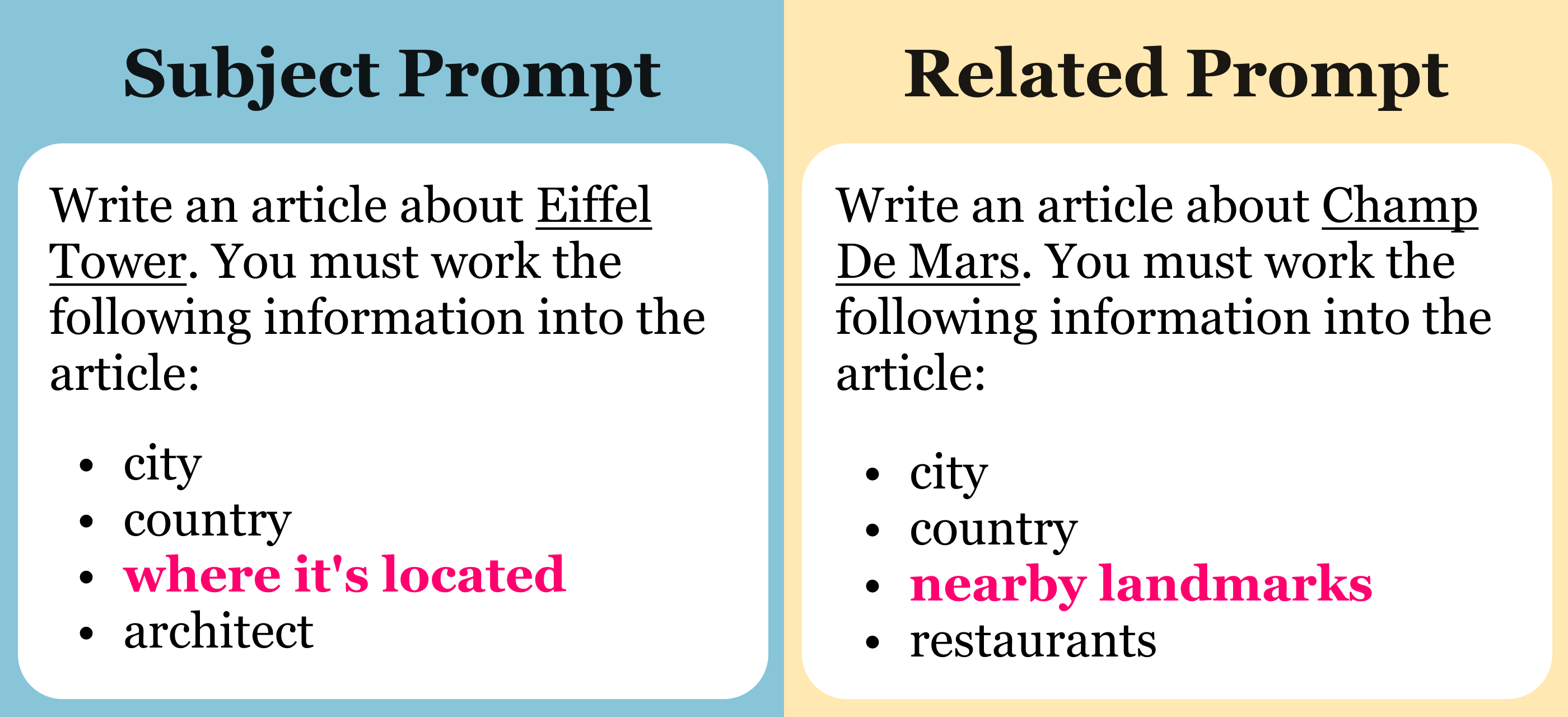}
    \caption{Example of prompts we used to generate passages to perform evaluation. The highlighted property means the subject (Champ De Mars or Eiffel Tower) is the object of that property (Where it's located or Nearby Landmarks). The edit for this example would be from "The Eiffel Tower is in Paris" to "The Eiffel Tower is in Rome"}.
    \label{fig:coupled-entity-prompts}
\end{figure}

Our dataset, \textbf{\textit{Coupled Entity Prompts}}, is based on zSRE \citep{levy_zero-shot_2017} and Counterfact \citep{meng_locating_2023}. We use the preparation of zSRE from \citet{mitchell_memory-based_2022, de_cao_editing_2021}, this dataset consists of factual statement and an alternative non-factual "edit" statement that would comprise an edit e.g. "What is the astronomical body that Ovda Regio is located on? Titan" and "What is the astronomical body that Ovda Regio is located on? Venus". Counterfact was originally developed by \citep{meng_locating_2023} to consist of edit statements that would previously be considered unlikely by a model before editing.

Each sample in our dataset consists of two prompts (please review \cref{fig:coupled-entity-prompts}) that will be used independently to generate paragraph-length outputs from LLMs. These prompts are highly coupled, where {\em coupling} is the degree to which a  subject entity ("Eiffel Tower") shares property with a related entity ("Champ De Mars") and the ground-truth target ("Paris"). In our example, Champ De Mars is the park the Eiffel Tower is located at. These entities are highly coupled since they share many properties such as city, country, and near by restaurants. Champ De Mars and Eiffel Tower both share the city of Paris as the ground-truth target which will be updated to Rome for model editing.

The \textit{subject prompt} asks the model to write an article about the subject of an edit (e.g., Eiffel Tower in \cref{fig:coupled-entity-prompts}) and to include a number of properties about that subject (e.g., where it's located). The \textit{related prompt} asks the model to write an article about a related entity (e.g., Champ De Mars in \cref{fig:coupled-entity-prompts}) and its properties where the related entity is highly coupled with the subject.

We define a successful edit in the ``long-form'' setting as: (1: \textit{Edit consistency}) completing the subject prompt as if the edit is true and the related passage does not contradict the edit, (2: \textit{Factual Consistency}) the subject and related passage minimize changes in the ground truth properties and (3: \textit{Internal Consistency}) the passages should neither contradict themselves nor each other.



We performed a SPARQL query on Wikidata to get related entities that had a relationship to both the subject and pre-edit target (e.g. Paris) for all subject entities in Counterfact and zSRE. We also queried for the ground truth properties about the subject and related entities (e.g. country, city, and restaurants near by). This data was used to construct prompts for a language model to write a paragraph about the subject and related entity and instructed the model to include those ground truth properties so we can measure portability and locality. In total, we constructed 3,867 subject and related entity prompts for Counterfact and 3,522 for zSRE (see \cref{app:dataset-consutrction} for details and examples). It's important to note each sample is accompanied by the original edit statement in order to measure the effect of an edit on the generated passages.

\subsection{Evaluation}

\paragraph{Likert Scale}
To measure the questions in \cref{sec:methods}, we devised a survey using a 7-point likert scale consisting of nine questions (subject and related passages were rated seperately; internal consistency includes a cross passage consistency rating). See \cref{app:survey-instrument} for full survey details. The survey is designed to assess the content that is generated as a result of the subject and related prompt. We call this content the subject and related passages respectively.

\paragraph{Human ratings}
\label{sec:method-evaluating-human-survey}
To collect human ratings, we randomly chose 12 samples from the Counterfact subset of our dataset. We use three methods to generate two outputs (subject and related passage) from the prompts in \cref{sec:coupled-entity-prompts-dataset} for each sample. First, we developed a \textbf{\textit{No edit}} control setting, where we used the language model \texttt{llama2-7b-chat} \citep{touvron_llama_2023} without making any edit intervention. These samples should rate low on edit consistency and act as a baseline for the other measures. Second, we used the editing method \textbf{\textit{ROME}} \citep{meng_locating_2023} to edit the model and then generate outputs. Finally, the authors of the paper wrote paragraph-length responses as if the edit were true to the same prompts to produce a \textbf{\textit{human}}-written baseline (see \cref{app:human-written-edit-details} for details). We expect the human-written baseline to score highest across all categories. The final result was 72 generated passages (\cref{app:model-details-generation} shows model generation details).

\paragraph{Sampling} The survey was distributed online to eight computer science graduate students who volunteered for the task from a research methods course. The study design included two groups of four participants. In each group, the participants rated the same randomly selected samples. The samples were in a random order to avoid fatigue bias. Each participant rated nine samples (three from each intervention) across the nine questions mentioned earlier. In total, we collected 648 ratings. Since each sample consists of two passages (the subject and related passage) and we had ratings for 144 passages. See \cref{app:survey-instrument} for full details. After adjudicating the survey results, we gathered survey results from four additional volunteer computer science graduate students to replace surveys that had very low agreement or poor quality (only rating the middle score for every answer).

\paragraph{Automatic Survey Ratings}
\label{sec:method-evaluating-automatic-survey}

The number of survey ratings was too small for a training set. So we generated synthetic ratings for generated passages resulting from 100 held-out coupled entity prompt samples.We generated the subject and related passages after performing the following model editing interventions: No edit, ROME \citep{meng_locating_2023}, IKE \citep{zheng_can_2023}, and FT \citep{zhu_modifying_2020}. We generated samples for \texttt{GPT-J} \citep{gpt-j-6b-a-8-billion} and \texttt{llama2-7b-chat}. We performed survey ratings using the same survey instructions human participants saw using \texttt{GPT-4} resulting in a total of 7,164 ratings (796 per question). Treating this as a training set, we trained \texttt{DeBERTav3} large \citep{he_debertav3_2022} for each question and evaluated the model using the human survey ratings as the test set. For experiments in \cref{sec:results-automatic-all-survey}, we train the models on the human survey ratings as well. See \cref{app:model-details-taking-a-survey} for details for an overview of how the model was trained and \cref{tab:automatic-agreement} for performance details.\footnote{We performed additional experiments with zero and few-shot settings using \texttt{GPT-3.5}, \texttt{GPT-4}, and \texttt{llama-2-7b-chat} (see \cref{app:model-details-taking-a-survey}).}

\subsection{Annotation}
\label{sec:method-evaluating-annotation}

In addition to a survey evaluation protocol, we presented annotators with a premise which consists of the subject or related entity passage and a claim which consists of one of the following: (1) an edit statement such as “The Eiffel Tower is located in Rome” (2) the pre-edit statement such as “The Eiffel Tower is located in Paris, or (3) a ground truth statement such as “The Eiffel Tower was completed in 1889”. Similar to natural language inference, each premise is classified as neutral to, supporting, or contradicting the claim. Additionally, annotators are instructed to highlight sentences that would provide evidence for the classification.

\label{sec:method-evaluating-human-annotation}
\paragraph{Human Annotations} Four authors of the paper performed annotations of 726 premise and claim pairs. We performed a pre-test before annotation, all four authors annotated a sample of 186 premise hypothesis pairs to understand the reliability of our annotation scheme. As measured by Krippendorff’s $\alpha$ the annotations had good agreement ($\alpha$ = 0.63). After the pre-test, the four authors split the remaining 540 annotations into two groups of 270 annotations and two annotators annotated each group ($\alpha$ = 0.65). In total, after adjudication for conflicts by a senior author (DR), there were 1,496 total classifications and 1,985 evidence sentences collected. See \cref{app:annotation-guidelines} for the annotation guidelines.

\paragraph{Automatic Annotations} \label{sec:method-evaluating-automatic-annotation} Since we have a large set of human annotations, we finetuned \texttt{DeBERTAv3} large \citep{he_debertav3_2022} on these. We enhanced the dataset with the highlighted sentences and treated those as premises for each claim resulting in a total of 1,642 samples after deduplication. To evaluate this method, we split the human annotations into a train test split of 80\% and 20\%\footnote{Only a single training run was performed without hyperparameter tuning so a validation split was not needed.}. See \cref{app:models-doing-annotation} for full training details and training set distribution. This appendix includes a comprehensive analysis of agreement scores.



\section{Experiments}
\label{sec:method-evaluating-automatic-full}

In order to answer our research question of how different model editing interventions compare, we develop a comprehensive suite of experiments that use the automatic measures developed above to evaluate the following interventions: FT with constraint loss \citep{zhu_modifying_2020}, MEND \citep{mitchell_fast_2022}, ROME \citep{meng_locating_2023}, MEMIT \citep{meng_mass-editing_2022}, and IKE \cite{zheng_can_2023}. 
We implemented these model editing interventions on \texttt{GPT2-XL} \citep{radford2019language}, \texttt{GPT-J} \citep{gpt-j-6b-a-8-billion}, \texttt{llama2-7b} and \texttt{llama2-7b-chat} \cite{touvron_llama_2023} (the main paper results report \texttt{GPT-J} and \texttt{llama2-7b-chat} with the other models results in \cref{app:additional-automatic-measures}). 
For each model, we also computed a `no edit' control which is simply using the coupled entity prompts to generate passages before performing any edit. We also experimented with using a zero shot \texttt{GPT-4} IKE setting (see \cref{app:gpt-4-ike}) to simulate an upper bound of performance. Subject and related prompts are completed as independent generations.


We perform these evaluations on 100 randomly sampled edits from Counterfact and zSRE. For the zSRE setting, we create two edits per sample. We compute a counterfactual edit, making an edit by changing a true fact to a counterfactual one, as well as a factual edit, changing a false fact to a true one. In total, we assess 300 samples (600 passages).

\begin{figure}[!t]
    \centering
    \includegraphics[width=\columnwidth]{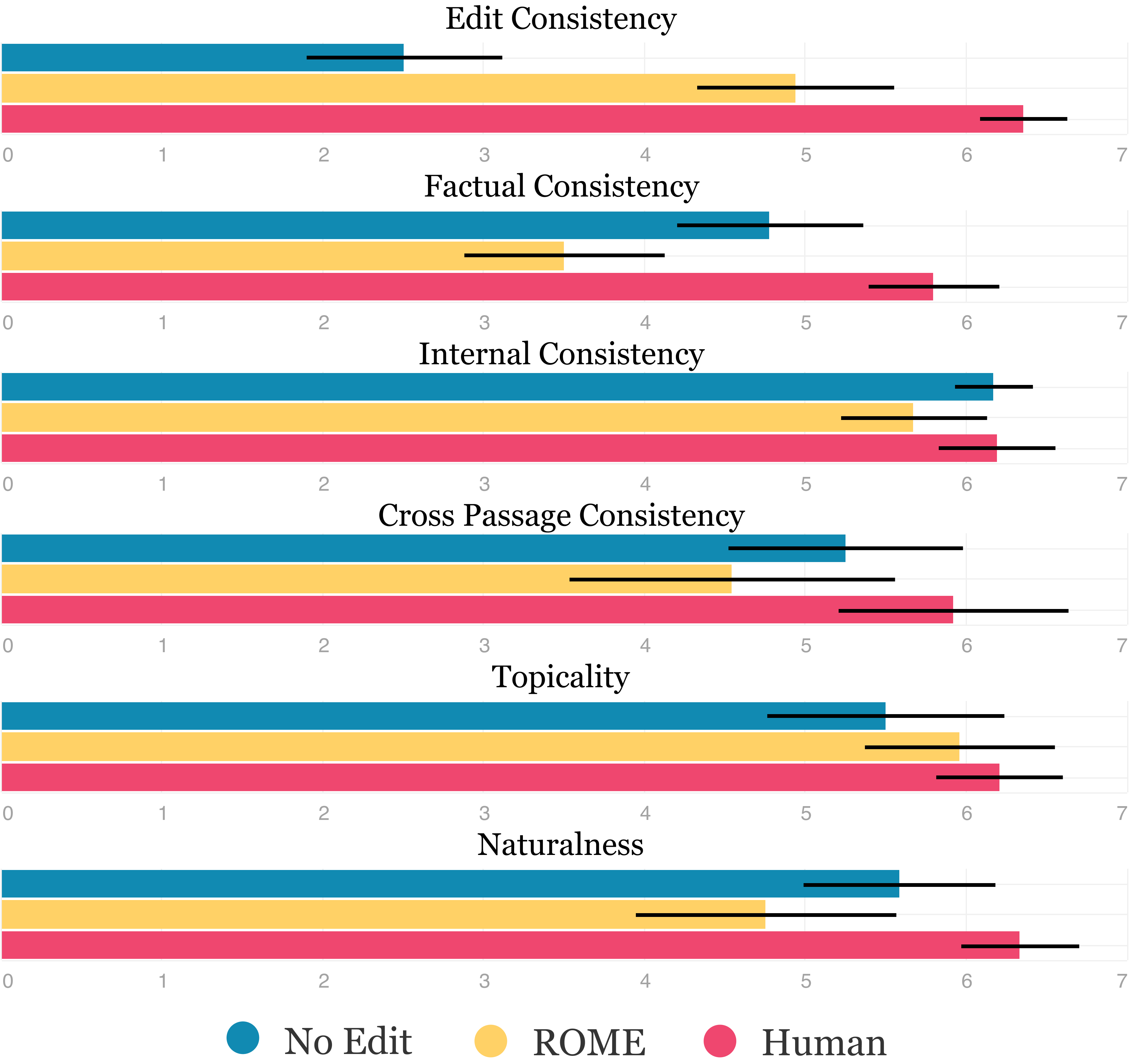}
    \caption{Survey results illustrating the mean rating of long-form quality measures. Human passages always rate highest. ROME is rated even worse than no edit on many dimensions.}
    \label{fig:pilot-survey-evaluation}
\end{figure}

\begin{figure*}[tp]
    \centering
    \includegraphics[scale=0.43]{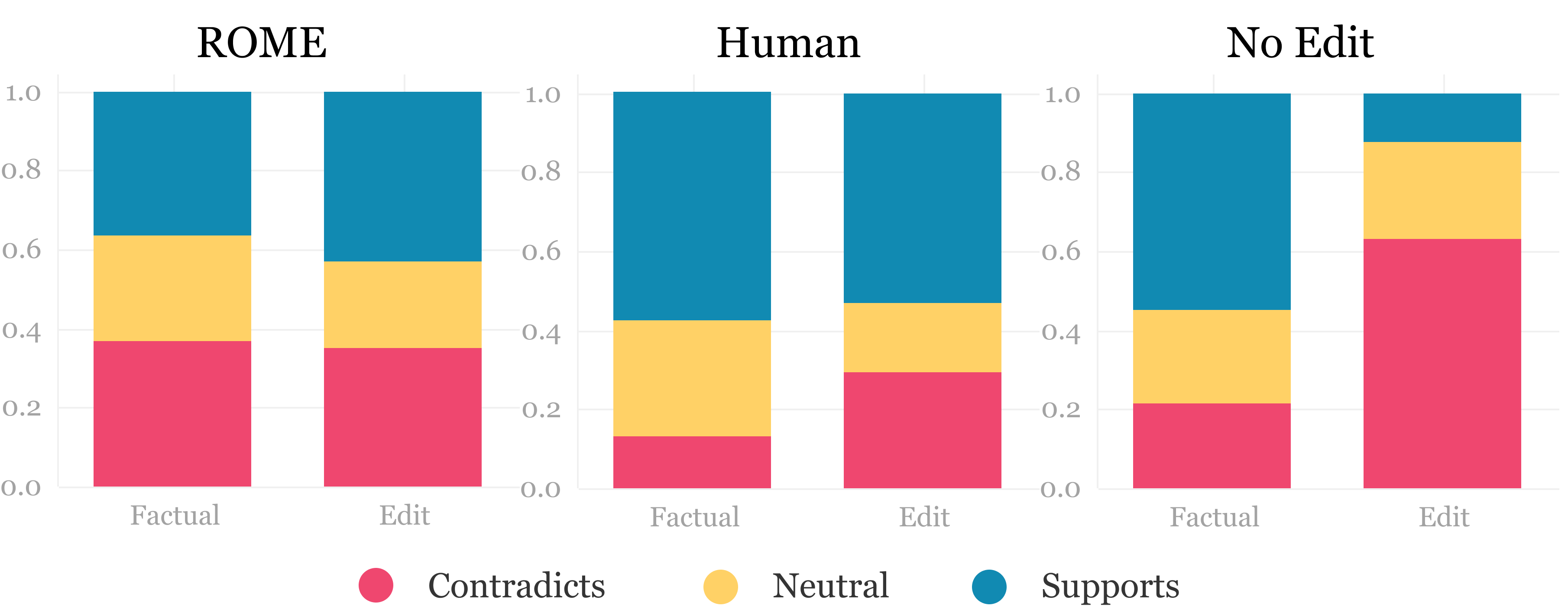}
    \caption{Proportion of labels from human annotation of ROME, human written, and no edit passages. The ground truth is mostly supported in the no edit and human control, while no edit mostly contradicts the edit statements. Human written passages generally are more consistent with the edit statement than ROME passages.}
    \label{fig:annotation-results}
\end{figure*}

\section{Results}
\label{sec:results}

Below we explore the results of our human evaluations as well as automatic evaluations.

\subsection{Human evaluation}
\label{sec:results-human-evaluation}

As we would expect (\cref{fig:pilot-survey-evaluation}), human written passages were rated higher than all other methods. ROME only approaches human ratings for edit consistency, internal consistency, and topicality. Interestingly the no edit control is rated higher than ROME in almost all dimensions except edit consistency and topicality. This indicates that ROME worsens the general quality of natural language generation. Cross passage consistency is reported separately from other internal consistency measures for illustrative purposes.
Ratings were statistically significant (one-sided Wilcoxon sign rank test, $p<.05$) except for no edit and human on internal and cross passage consistency and human and ROME on topicality.

For the annotations, \cref{fig:annotation-results} corroborates our survey findings: both the no edit and human control groups have better factual consistency than ROME as measured by the number of ground truth statements that are supported. Human written passages have better factual consistency and edit consistency than ROME or no edit. All comparisons in between methods were statistically significant (Chi-square test of independence). See \cref{tab:annotation-label-distribution} for annotation distribution details.

\begin{table*}[!t]
\small
\begin{tabular}{lllllllllll}
\toprule
    Model              &  Method       & \multicolumn{2}{l}{\textbf{Edit consistency}} & \multicolumn{2}{l}{\textbf{Factual consistency}} & \multicolumn{3}{l}{\textbf{Internal consistency}}               & \textbf{Topicality}       & \textbf{Naturalness}      \\ 
                &        & Subject              & Related          & Subject               & Related            & Subject             & Related          & Cross            &                  &                  \\ \midrule
\texttt{GPT-J}  & No Edit       & 1.3\tiny{$\pm$1.3}          & 3.5\tiny{$\pm$1.3}          & {\ul 2.3\tiny{$\pm$1.8}}       & {\ul 3.8\tiny{$\pm$2.3}}       & \textbf{6.6\tiny{$\pm$1.5}}     & \textbf{7.0\tiny{$\pm$0.1}}     & \textbf{6.6\tiny{$\pm$1.0}}          & 5.4\tiny{$\pm$2.3}                                                                    & \textbf{5.4\tiny{$\pm$2.6}}                                                                                    \\
                & IKE          & 2.0\tiny{$\pm$2.2}          & {\ul 3.9\tiny{$\pm$1.6}}    & \textbf{2.4\tiny{$\pm$1.7}}    & \textbf{4.1\tiny{$\pm$2.3}}    & 6.4\tiny{$\pm$1.7}              & {\ul 6.9\tiny{$\pm$0.4}}        & {\ul 6.5\tiny{$\pm$1.0}}             & 5.4\tiny{$\pm$2.1}                                                                    & {\ul 5.3\tiny{$\pm$2.6}}                                                                                       \\
                & FT           & 1.5\tiny{$\pm$1.7}          & 3.8\tiny{$\pm$1.1}          & 2.1\tiny{$\pm$1.6}             & \textbf{4.1\tiny{$\pm$2.3}}    & {\ul 6.5\tiny{$\pm$1.5}}        & {\ul 6.9\tiny{$\pm$0.7}}        & \textbf{6.6\tiny{$\pm$1.0}}          & \textbf{5.6\tiny{$\pm$2.1}}                                                           & \textbf{5.4\tiny{$\pm$2.6}}                                                                                    \\
                & MEND         & \textbf{3.2\tiny{$\pm$2.9}} & \textbf{4.0\tiny{$\pm$1.8}} & \textbf{2.4\tiny{$\pm$1.8}}    & \textbf{4.1\tiny{$\pm$2.4}}    & 6.0\tiny{$\pm$2.2}              & 6.7\tiny{$\pm$1.3}              & {\ul 6.5\tiny{$\pm$1.2}}             & 4.5\tiny{$\pm$2.5}                                                                    & 5.1\tiny{$\pm$2.7}                                                                                             \\
                & ROME         & {\ul 2.8\tiny{$\pm$2.8}}    & {\ul 3.9\tiny{$\pm$1.4}}    & 1.5\tiny{$\pm$1.1}             & 3.2\tiny{$\pm$2.2}             & 5.9\tiny{$\pm$2.1}              & {\ul 6.9\tiny{$\pm$0.6}}        & 6.0\tiny{$\pm$1.4}                   & 4.1\tiny{$\pm$2.5}                                                                    & 4.4\tiny{$\pm$2.9}                                                                                             \\
                & MEMIT        & 2.1\tiny{$\pm$2.3}          & 3.8\tiny{$\pm$1.3}          & 2.0\tiny{$\pm$1.6}             & 3.7\tiny{$\pm$2.2}             & {\ul 6.5\tiny{$\pm$1.6}}        & {\ul 6.9\tiny{$\pm$0.5}}        & {\ul 6.5\tiny{$\pm$1.1}}             & {\ul 5.5\tiny{$\pm$2.1}}                                                              & {\ul 5.3\tiny{$\pm$2.7}}                                                                                       \\ \midrule
\texttt{llama2} & No Edit       & 2.2\tiny{$\pm$2.4}          & 2.0\tiny{$\pm$1.5}          & \textbf{3.5\tiny{$\pm$1.9}}    & {\ul 4.7\tiny{$\pm$2.4}}       & \textbf{6.9\tiny{$\pm$0.3}}     & \textbf{7.0\tiny{$\pm$0.1}}     & \textbf{6.6\tiny{$\pm$1.4}}          & \textbf{7.0\tiny{$\pm$0.4}}                                                           & \textbf{6.9\tiny{$\pm$0.8}}                                                                                    \\
                & IKE          & 4.7\tiny{$\pm$2.9}          & 3.4\tiny{$\pm$2.4}          & {\ul 3.4\tiny{$\pm$1.9}}       & \textbf{4.9\tiny{$\pm$2.2}}    & {\ul 6.8\tiny{$\pm$0.6}}        & \textbf{7.0\tiny{$\pm$0.2}}     & \textbf{6.6\tiny{$\pm$1.4}}          & \textbf{7.0\tiny{$\pm$0.2}}                                                           & {\ul 6.8\tiny{$\pm$1.0}}                                                                                       \\
                & FT           & {\ul 5.1\tiny{$\pm$2.8}}    & \textbf{3.7\tiny{$\pm$2.4}} & 2.1\tiny{$\pm$1.5}             & 3.7\tiny{$\pm$2.4}             & 5.7\tiny{$\pm$2.4}              & 6.7\tiny{$\pm$1.4}              & {\ul 6.1\tiny{$\pm$1.6}}             & 5.1\tiny{$\pm$2.7}                                                                    & 5.8\tiny{$\pm$2.4}                                                                                             \\
                & MEND         & 3.1\tiny{$\pm$2.9}          & 2.6\tiny{$\pm$1.9}          & {\ul 3.4\tiny{$\pm$1.9}}       & 4.6\tiny{$\pm$2.3}             & {\ul 6.8\tiny{$\pm$0.6}}        & \textbf{7.0\tiny{$\pm$0.1}}     & \textbf{6.6\tiny{$\pm$1.3}}          & {\ul 6.9\tiny{$\pm$0.5}}                                                              & {\ul 6.8\tiny{$\pm$1.2}}                                                                                       \\
                & ROME         & \textbf{5.4\tiny{$\pm$2.6}} & {\ul 3.5\tiny{$\pm$2.4}}    & 1.9\tiny{$\pm$1.4}             & 3.9\tiny{$\pm$2.4}             & 6.4\tiny{$\pm$1.7}              & \textbf{7.0\tiny{$\pm$0.5}}     & 5.8\tiny{$\pm$2.1}                   & 6.5\tiny{$\pm$1.5}                                                                    & 6.2\tiny{$\pm$2.0}                                                                                             \\
                & MEMIT        & \textbf{5.4\tiny{$\pm$2.7}} & 3.3\tiny{$\pm$2.3}          & 2.0\tiny{$\pm$1.5}             & 3.8\tiny{$\pm$2.4}             & 6.3\tiny{$\pm$1.8}              & {\ul 6.9\tiny{$\pm$0.6}}        & 5.9\tiny{$\pm$2.1}                   & 6.3\tiny{$\pm$1.8}                                                                    & 6.2\tiny{$\pm$2.0}                                                                                             \\ \midrule
\texttt{GPT-4}           & IKE          & 5.2\tiny{$\pm$2.7}          & 5.1\tiny{$\pm$2.5}          & 3.2\tiny{$\pm$1.9}             & 6.1\tiny{$\pm$1.5}             & 6.7\tiny{$\pm$1.3}              & 7.0\tiny{$\pm$0.0}              & 6.7\tiny{$\pm$1.2}                   & 6.7\tiny{$\pm$1.4}                                                                    & 7.0\tiny{$\pm$0.0}                                                                                            \\                     
\bottomrule
\end{tabular}
\caption{Automatic ratings of zSRE and Counterfact (\texttt{DeBERTaV3}) across editing methods. Significant reduction (one-sided Wilcoxon sign rank, $p<0.05$) in factual consistency for ROME and MEMIT. \texttt{llama2} here is \texttt{llama2-7b-chat}}.
    \label{tab:automatic-ratings}
\end{table*}

\subsection{Understanding the impact of model editing across interventions}
\label{sec:results-automatic-all-survey}


\cref{tab:automatic-ratings} illustrates the quality of various model editing methods using our automatic survey rating approach. Our main findings is that ROME and MEMIT suffer from significant drops in performance on factual consistency\footnote{We found no statistically significant correlation between factual consistency and edit consistency.} and internal consistency (especially cross passage) despite often being the most effective editing method according to short-form evaluations (\cref{app:short-evals}). Except for \texttt{GPT-4} IKE, there seems to be a pattern where models that do better at edit consistency for the subject passages perform worse on reflecting the edit in the related passages. Unsurprisingly in-context editing (IKE) tends to maintain similar performance to the `no edit' control across factual consistency, internal consistency, topicality, and naturalness despite not being as effective at edit consistency. Along these lines, \texttt{GPT-4} IKE is generally the most effective method especially in the case of maintaining edit consistency and factual consistency in the related passages\footnote{Additional comparisons with other models and automatic measures such as zero-shot and simpler baselines are presented in \cref{app:additional-automatic-measures}.}.




\subsection{What is the scope of the edit?}
\label{sec:results-automatic-classification}

\begin{figure*}[t]
    \centering
    \includegraphics[scale=0.3]{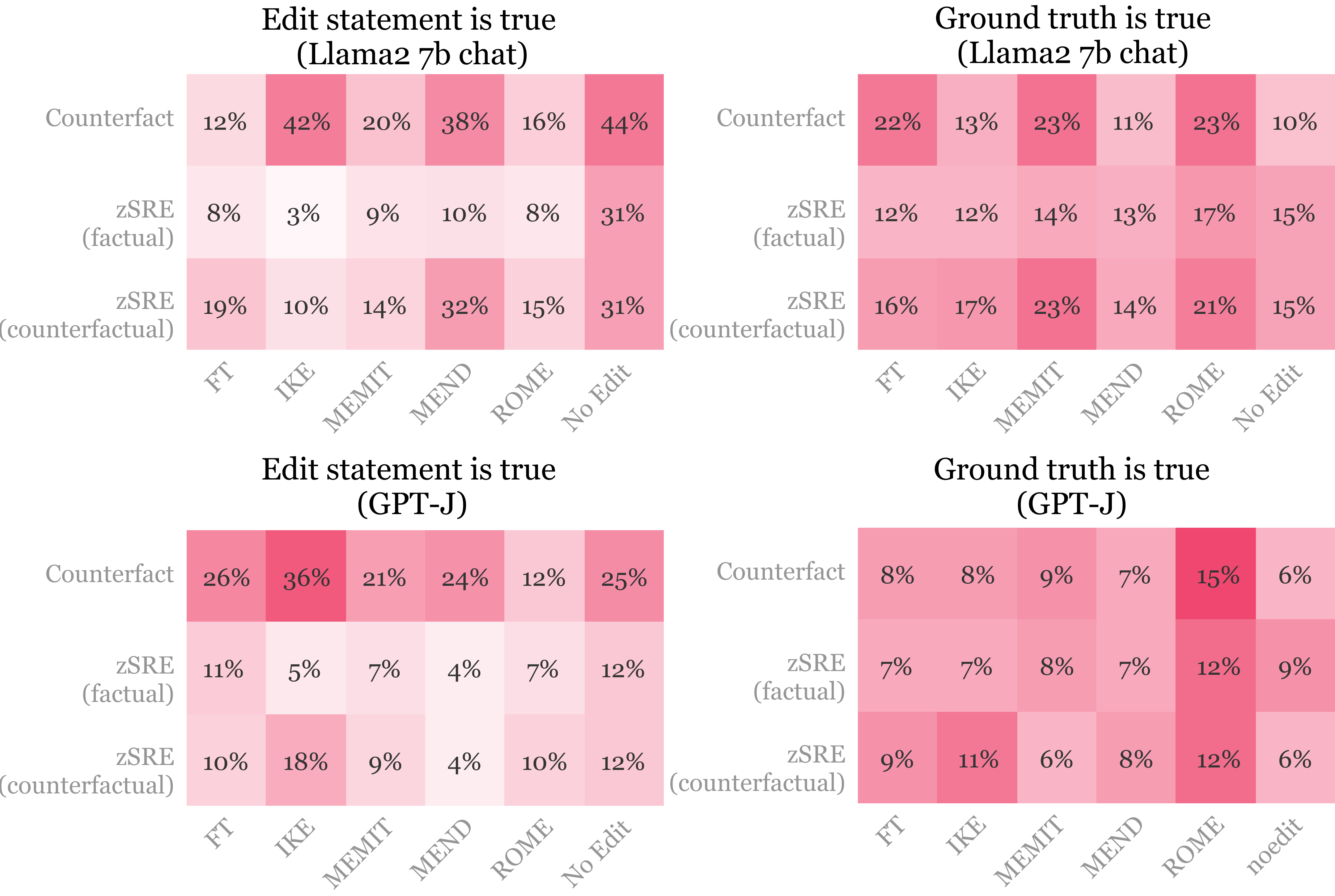}
    \caption{Percentage of claims that contradict the generated passage. Results corroborate our findings that MEMIT and ROME suffer from high factual drift.}
    \label{fig:classifications-results}
\end{figure*}

Our classifier allows us to \textit{understand the scope of the change introduced by an editing method} since we are able to measure the number of ground truth properties that are contradicted by the generated passage after the edit (see \cref{fig:classifications-results}). Importantly, no edit indicates the base level of ground truth or edit statements that would be contradicted before the edit was made. All methods perform better than the no edit control on ensuring the edit statement is not contradicted with particular effectiveness of MEMIT, ROME, and FT on \texttt{lama2-7b-chat}. However, MEMIT and ROME introduce a high degree of ``factual drift'' (suffer from locality problems) since a higher \% of ground truth statements are contradicted compared to the no edit control and the other methods. we found no inverse relationship between edit and factual consistency.


\subsection{Correlating long- and short-form evaluations?}
\label{sec:results-long-short-relationship}
We only found very weak relationships between the short-form evaluations of edit success, generalization, locality, and portability settings from \citet{yao_editing_2023} and long-form evaluations (\cref{tab:correlation-with-long-form}). Edit consistency generally does capture some of what is measured by the short-form metrics ($\rho \in [0.1,0.17]$, $p<0.05$). Cross passage consistency also has weak correlations with portability ($\rho = 0.13$, $p<0.05$) and generalization ($\rho = 0.12$, $p<0.05$). Importantly, factual consistency and internal consistency have almost no relationship with short-form measures ($\rho \in [-0.08,0.05]$, $p<0.05$). We speculate the reason for this is that the short-form metrics measure superficial token distribution questions about word co-occurrence (see \citet{hoelscher-obermaier-etal-2023-detecting} for an illustration) while our measures require success across much larger generations. Either way, this finding indicates that our evaluation setting measures unique dimensions not captured by short-form evaluation.

\subsection{Injection vs updating facts}
\label{sec:results-injection-v-update}

\begin{figure}[h]
    \centering
    \includegraphics[width=\columnwidth]{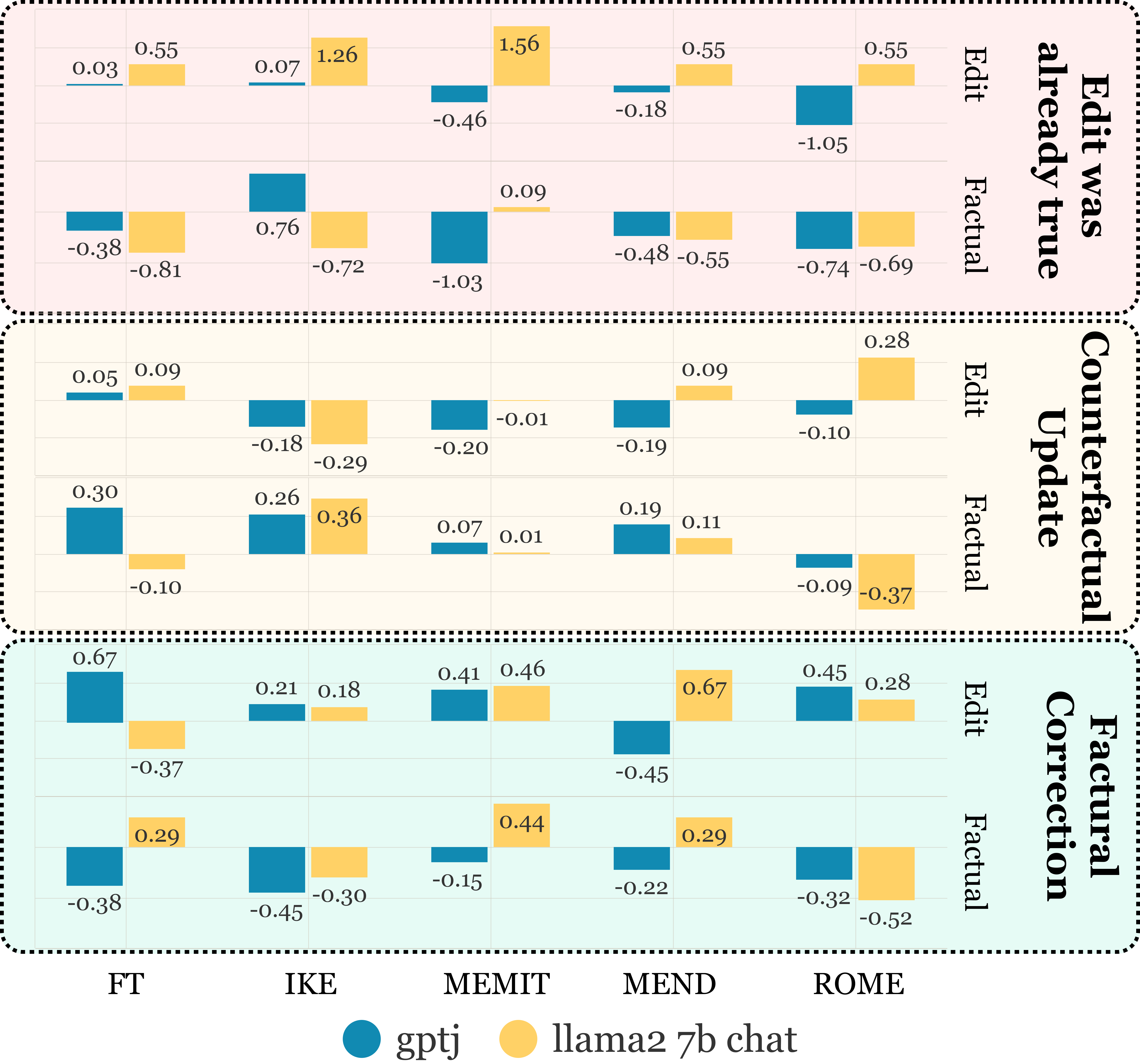}
    \caption{Model performance can differ depending on the type of edit task.}
    \label{fig:performance-analysis}
\end{figure}

One limitation with model editing evaluations is that we are not sure if we are updating a previously known fact or injecting a brand new fact since knowing a fact beforehand is model specific\footnote{See a similar analysis in 5.1 of \citep{hase2023does}}. We analyze the performance difference between these in \cref{fig:performance-analysis} by looking at the mean rating difference on edit consistency and factual consistency measures considering whether \textit{an edit statement was already known} or not (\textbf{Edit was already true}), whether the edit is a \textit{counterfactual update} or a novel fact injection (\textbf{Counterfactual update}) and whether the edit is \textit{factual correction} of a known but wrong fact or is a novel fact injection. \cref{tab:proportion-statistics} illustrates the proportion of samples that represent these categories for each dataset.

We see a small performance drop on edit consistency if we are doing a counterfactual update rather than a novel fact injection indicating updates are harder than injection (updates only represent 8\% and 18\% of Counterfact on \texttt{GPT-J} and \texttt{llama2-7b-chat} see \cref{tab:proportion-statistics}). Factual consistency is better for counterfactual updates compared to novel fact injection which might mean that during a novel fact injection, we are simply missing additional necessary ground truth knowledge. For factual correction, we see that generally we do better on edit consistency if we are correcting an erroneous fact. For factual consistency, we do worse in the factual correction setting with some exceptions which means that novel fact injection is easier to maintain ground truth statements on when a model already is biased towards and incorrect answer.

Finally, \cref{tab:proportion-statistics} shows how the edit statement is already true in many cases in zSRE. In \cref{fig:performance-analysis} we see the implications of this where for \texttt{llama2-7b-chat}, if the edit was already true then edit consistency is rated much higher and, as we’d expect since this is a statement that would contradict ground truth, factual consistency is much lower. Overall, these differences aren't large enough to change our results in \cref{sec:results-automatic-all-survey} but we should perform these types of controlled experiments when doing model editing experiments to ensure our results hold across different types of editing tasks\footnote{For the readers benefit we present a similar performance analysis for the short evaluations in \cref{app:short-evals-performance}}.

\subsection{Error Analysis}
\label{sec:results-error-analysis}

In order to understand particular errors made during generation, we manually analyzed 200 samples from Counterfact by selecting the 20 lowest automatically rated samples for each edit intervention for \texttt{GPT-J} and \texttt{llama2-7b-chat}. Due to space limitations, please reference the qualitative examples in \cref{app:examples} during the discussion.

First, we found a number of cases of disfluency. Aside from common cases of disfluency in NLG like repetition or completely degenerate generations (often from FT), we found there were cases of nonsensical generations like in Example \hyperref[example:1]{1} where Boston gets overused as a noun for categories like profession. Example \hyperref[example:2]{2} illustrates a relatively common degenerative case with ROME and MEMIT where space tokens were omitted.

Another common problem was cases with entity or topic drift and lexical cohesion issues. In Example \hyperref[example:3]{3} MEMIT correctly edits Paul Guimard’s birth place to be in Russia but the change creates a whole new entity with the same name who is a Russian cosplayer born in the 1980s, the Paul Guimard \textit{we intended} to edit stays unedited as reflected in the related passage. Examples \hyperref[example:4]{4} and \hyperref[example:8]{8} illustrate a common case where the subject entity is introduced at the beginning but the generated passage slowly drifts towards another entity (in this case the Empire Building or IBM Lotus) and continues to drift into another topic. Examples \hyperref[example:5]{5} and \hyperref[example:7]{7} illustrate cases of poor lexical cohesion where the name of the entity slowly changes over the course of the generation (e.g. Delon becomes Deloy which becomes Deloyg). Another illustrative example from a ROME edit in the human survey is 
\begin{footnotesize}
[Benedetto Marcello (1847-1937) was an Italian jazz musician... He was born in Genoa, Italy, to parents Antonino and Teresa Jazz. His family name is Benedetto Jazz]
\end{footnotesize}
Example \hyperref[example:6]{6} is a combination of topic and entity drift where Milan is correctly edited to be located in Japan but the generation drifts towards talking about Milan as if it were an alias for Tokyo and continues referring to the subject as Tokyo rather than Milan.

Another common case of editing failure is the introduction of contradictions that either contradict with statements made during the main passage (within a single generation) or that conflict with other generations in the related passage. Example \hyperref[example:10]{10} states the Ipod was created by Nintendo and then in the next sentence mentions it was created by Apple. Example \hyperref[example:13]{13} mentions that Guimard was Groult’s cousin but in the related passage they are said to be married. Other edits contradict common sense or world knowledge such as Example \hyperref[example:16]{16} where the Dawa River is a river located in Malta but later mentions how the Dawa is a tributary of the Jubba River which the model says is in Somalia.

Finally, reflecting our finding that some models tend to violate more ground truth properties than others, we found success cases where some models only made minimal edits (Example \hyperref[example:17]{17}) or edits that incorporate both the edit statement and the pre-edited fact (Example \hyperref[example:11]{11}), while other edits introduced very large changes violating locality such as Example \hyperref[example:14]{14} where changing Jeanne Moreau's birth place to Poland unnecessarily changes her teacher Denis d'Inès to be Polish as well when generating the related passage (a reflection of poor locality). Again Example \hyperref[example:15]{15} does not just change the band Barren Earth’s location to Sydney, Australia but also changes the subgenre of the band as well as the members of the band. While IKE is generally an effective method for editing larger models can reject the edit. Example \hyperref[example:20]{20} illustrates a case with \texttt{GPT-4} IKE where the edit is rejected by the model.


\section{Discussion}
\label{sec:discussion}

Current model editing methods have many gaps that are not measured by short-form evaluation methods and the preliminary `long-form` methods from \citep{meng_locating_2023} don't correlate well with human data. Factual drift, where methods like ROME tend to make much larger changes than MEND or FT is not revealed in the standardized `shot-form` measures from \citet{yao_editing_2023}. Future efforts should be devoted to balancing factual drift and edit success in NLG.


Factual drift might be a desirable feature of model editing, where there are model edits that should imply changes that would contradict ground truth statements. However, we want to develop evaluation methods that are able to measure the trade off between edit and factual consistency which we believe our methods are able to measure. RippleEdit \citep{cohen_evaluating_2023} is a good step in this direction for short-form evaluation that could inspire future work on more comprehensive long form evaluations that measure the scope of change beyond a single related passage.


Finally, our results reveal an important general property we should be looking for in high-quality model editing methods: consistency. The problematic generations that we investigated often indicated cases of contradiction, whether that was self contradiction, contradicting separated generations in the related passages, or contradicting ground truth statements. As developers of model editing interventions, we should design methods that result in generations that have high consistency: there should at least be no contradictions across generated passages. For cases where we allow a high factual drift, we still want to ensure self consistency. Other properties like fluency and topicality are important properties which tend to suffer and we should ensure that novel methods do not inadvertently harm general NLG quality.

\section{Conclusion}

In this paper, we introduced two automatic methods, survey ratings and classification, for evaluating the impact of model editing on natural language generation in paragraph-length generation settings. We validated these measures by collecting survey and annotation data from human participants and then developed a trained model setting that correlated well with human data.

Using these automatic metrics, we performed a comprehensive analysis of the natural language generation quality of common model editing techniques finding the following results: (1) ROME and MEMIT suffer from a high factual drift from ground truth statements compared to other methods like MEND or IKE (2) there is very little relationship between previous short form evaluations like generalization, locality, and portability with our long form metrics (3) through a qualitative study, we presented a number of common failure modes such as entity drift, lexical cohesion, internal contradiction, and scope errors. We hope that identifying these failure modes can help the community develop future model editing techniques that work well in ``long-form'' settings.


\section{Limitations}

The primary limitation of our study is the small sample size and weak inter-rater reliability of our survey filled out by human participants. It is important to note that we initially had a Krippendorff $\alpha$ of .3 and reran the survey with a new of set of participants but only marginally increased the agreement to .34\footnote{Agreement on some measures like edit consistency were much higher ($\alpha=0.55$) see \cref{tab:inter-rater-reliability}.}. The surveys took on average one hour to complete and is much more laborious to complete than the annotations. To further develop the survey method we should investigate ways of increasing both the inter-rater reliability and efficiency.

We performed the study using a limited demographic of graduate computer science students who would be familiar with the language of natural language generation. Studies looking to scale up our method with diverse demographics such as from crowdsourcing would likely suffer from even worse agreement. One alternative could be finding alternative ways to operationalize measures like internal consistency and topicality.

Another limitation is that our methods only implicitly captures generalization, locality, and portability so we can’t speak directly to specific effects on these properties with our measure. Related, the study only uses one related entity when generating and assessing our related passage. To further assess the scope of impact, future methods should incorporate generated passages farther away in the knowledge graph than the highest coupled entities.

One notable gap in our study that should be followed up on is the question of the impact of batch and chained editing has on NLG quality. Since we can imagine many settings in which a user would want to make a large amount of edits to a language model or make subsequent edits one after another, we would want to understand what impact that has on NLG separately from short evaluations.

Finally, it is important to acknowledge the ethical concerns with using counterfactual editing datasets. These datasets purposely introduce misinformation to determine the efficacy of an editing technique. As a community we should be aware that a side effect of this research could be demonstrating comprehensive methods for injecting misinformation and as such we should look towards moving away from counterfactual editing towards factual correction datasets or datasets that have less misinformation harm risk such as edits in fictional settings.

\section*{Acknowledgements}

We thank the Digital Research Alliance of Canada and Vector Institute for the use of compute resources. DR's work is supported by the Killam Foundation through the Killam Predoctoral Fellowship. We also acknowledge the guidance of Lizbeth Olivia Escobedo Bravo and help from student volunteers from her Research Methods course.

\bibliography{custom}

\begin{thebibliography}{23}
\expandafter\ifx\csname natexlab\endcsname\relax\def\natexlab#1{#1}\fi

\bibitem[{Cohen et~al.(2023)Cohen, Biran, Yoran, Globerson, and Geva}]{cohen_evaluating_2023}
Roi Cohen, Eden Biran, Ori Yoran, Amir Globerson, and Mor Geva. 2023.
\newblock \href {https://doi.org/10.48550/arXiv.2307.12976} {Evaluating the {Ripple} {Effects} of {Knowledge} {Editing} in {Language} {Models}}.
\newblock ArXiv:2307.12976 [cs].

\bibitem[{De~Cao et~al.(2021)De~Cao, Aziz, and Titov}]{de_cao_editing_2021}
Nicola De~Cao, Wilker Aziz, and Ivan Titov. 2021.
\newblock \href {https://doi.org/10.18653/v1/2021.emnlp-main.522} {Editing {Factual} {Knowledge} in {Language} {Models}}.
\newblock \emph{Proceedings of the 2021 Conference on Empirical Methods in Natural Language Processing}, pages 6491--6506.
\newblock Conference Name: Proceedings of the 2021 Conference on Empirical Methods in Natural Language Processing Place: Online and Punta Cana, Dominican Republic Publisher: Association for Computational Linguistics.

\bibitem[{Hase et~al.(2023)Hase, Bansal, Kim, and Ghandeharioun}]{hase2023does}
Peter Hase, Mohit Bansal, Been Kim, and Asma Ghandeharioun. 2023.
\newblock \href {http://arxiv.org/abs/2301.04213} {Does localization inform editing? surprising differences in causality-based localization vs. knowledge editing in language models}.

\bibitem[{Hase et~al.(2021)Hase, Diab, Celikyilmaz, Li, Kozareva, Stoyanov, Bansal, and Iyer}]{hase_language_2021}
Peter Hase, Mona Diab, Asli Celikyilmaz, Xian Li, Zornitsa Kozareva, Veselin Stoyanov, Mohit Bansal, and Srinivasan Iyer. 2021.
\newblock \href {http://arxiv.org/abs/2111.13654} {Do {Language} {Models} {Have} {Beliefs}? {Methods} for {Detecting}, {Updating}, and {Visualizing} {Model} {Beliefs}}.
\newblock ArXiv:2111.13654 [cs].

\bibitem[{He et~al.(2022)He, Gao, and Chen}]{he_debertav3_2022}
Pengcheng He, Jianfeng Gao, and Weizhu Chen. 2022.
\newblock \href {https://openreview.net/forum?id=sE7-XhLxHA} {{DeBERTaV3}: {Improving} {DeBERTa} using {ELECTRA}-{Style} {Pre}-{Training} with {Gradient}-{Disentangled} {Embedding} {Sharing}}.

\bibitem[{Hernandez et~al.(2023)Hernandez, Li, and Andreas}]{hernandez_inspecting_2023}
Evan Hernandez, Belinda~Z. Li, and Jacob Andreas. 2023.
\newblock \href {https://www.semanticscholar.org/paper/Inspecting-and-Editing-Knowledge-Representations-in-Hernandez-Li/a206a0c96d6076c6ab081288b0c2c95d3c7efd64} {Inspecting and {Editing} {Knowledge} {Representations} in {Language} {Models}}.

\bibitem[{Hoelscher-Obermaier et~al.(2023)Hoelscher-Obermaier, Persson, Kran, Konstas, and Barez}]{hoelscher-obermaier-etal-2023-detecting}
Jason Hoelscher-Obermaier, Julia Persson, Esben Kran, Ioannis Konstas, and Fazl Barez. 2023.
\newblock \href {https://doi.org/10.18653/v1/2023.findings-acl.733} {Detecting edit failures in large language models: An improved specificity benchmark}.
\newblock In \emph{Findings of the Association for Computational Linguistics: ACL 2023}, pages 11548--11559, Toronto, Canada. Association for Computational Linguistics.

\bibitem[{Huang et~al.(2023)Huang, Shen, Zhang, Zhou, Rong, and Xiong}]{huang_transformer-patcher_2023}
Zeyu Huang, Yikang Shen, Xiaofeng Zhang, Jie Zhou, Wenge Rong, and Zhang Xiong. 2023.
\newblock \href {https://doi.org/10.48550/ARXIV.2301.09785} {Transformer-{Patcher}: {One} {Mistake} worth {One} {Neuron}}.
\newblock Publisher: arXiv Version Number: 1.

\bibitem[{Laurer et~al.(2022)Laurer, van Atteveldt, Casas, and Welbers}]{laurer2022less}
Moritz Laurer, Wouter van Atteveldt, Andreu Casas, and Kasper Welbers. 2022.
\newblock Less annotating, more classifying: Addressing the data scarcity issue of supervised machine learning with deep transfer learning and bert-nli.
\newblock \emph{Political Analysis}, pages 1--33.

\bibitem[{Levy et~al.(2017)Levy, Seo, Choi, and Zettlemoyer}]{levy_zero-shot_2017}
Omer Levy, Minjoon Seo, Eunsol Choi, and Luke Zettlemoyer. 2017.
\newblock \href {https://doi.org/10.18653/v1/K17-1034} {Zero-{Shot} {Relation} {Extraction} via {Reading} {Comprehension}}.
\newblock In \emph{Proceedings of the 21st {Conference} on {Computational} {Natural} {Language} {Learning} ({CoNLL} 2017)}, pages 333--342, Vancouver, Canada. Association for Computational Linguistics.

\bibitem[{Meng et~al.(2023)Meng, Bau, Andonian, and Belinkov}]{meng_locating_2023}
Kevin Meng, David Bau, Alex Andonian, and Yonatan Belinkov. 2023.
\newblock \href {http://arxiv.org/abs/2202.05262} {Locating and {Editing} {Factual} {Associations} in {GPT}}.
\newblock ArXiv:2202.05262 [cs].

\bibitem[{Meng et~al.(2022)Meng, Sharma, Andonian, Belinkov, and Bau}]{meng_mass-editing_2022}
Kevin Meng, Arnab~Sen Sharma, Alex~J. Andonian, Yonatan Belinkov, and David Bau. 2022.
\newblock \href {https://openreview.net/forum?id=MkbcAHIYgyS} {Mass-{Editing} {Memory} in a {Transformer}}.

\bibitem[{Mitchell et~al.(2022{\natexlab{a}})Mitchell, Lin, Bosselut, Finn, and Manning}]{mitchell_fast_2022}
Eric Mitchell, Charles Lin, Antoine Bosselut, Chelsea Finn, and Christopher~D. Manning. 2022{\natexlab{a}}.
\newblock \href {http://arxiv.org/abs/2110.11309} {Fast {Model} {Editing} at {Scale}}.
\newblock ArXiv:2110.11309 [cs].

\bibitem[{Mitchell et~al.(2022{\natexlab{b}})Mitchell, Lin, Bosselut, Manning, and Finn}]{mitchell_memory-based_2022}
Eric Mitchell, Charles Lin, Antoine Bosselut, Christopher~D. Manning, and Chelsea Finn. 2022{\natexlab{b}}.
\newblock \href {https://doi.org/10.48550/arXiv.2206.06520} {Memory-{Based} {Model} {Editing} at {Scale}}.
\newblock ArXiv:2206.06520 [cs].

\bibitem[{Radford et~al.(2019)Radford, Wu, Child, Luan, Amodei, Sutskever et~al.}]{radford2019language}
Alec Radford, Jeffrey Wu, Rewon Child, David Luan, Dario Amodei, Ilya Sutskever, et~al. 2019.
\newblock Language models are unsupervised multitask learners.
\newblock \emph{OpenAI blog}, 1(8):9.

\bibitem[{Thorne et~al.(2018)Thorne, Vlachos, Christodoulopoulos, and Mittal}]{thorne2018fever}
James Thorne, Andreas Vlachos, Christos Christodoulopoulos, and Arpit Mittal. 2018.
\newblock Fever: a large-scale dataset for fact extraction and verification.
\newblock \emph{arXiv preprint arXiv:1803.05355}.

\bibitem[{Touvron et~al.(2023)Touvron, Martin, Stone, Albert, Almahairi, Babaei, Bashlykov, Batra, Bhargava, Bhosale, Bikel, Blecher, Ferrer, Chen, Cucurull, Esiobu, Fernandes, Fu, Fu, Fuller, Gao, Goswami, Goyal, Hartshorn, Hosseini, Hou, Inan, Kardas, Kerkez, Khabsa, Kloumann, Korenev, Koura, Lachaux, Lavril, Lee, Liskovich, Lu, Mao, Martinet, Mihaylov, Mishra, Molybog, Nie, Poulton, Reizenstein, Rungta, Saladi, Schelten, Silva, Smith, Subramanian, Tan, Tang, Taylor, Williams, Kuan, Xu, Yan, Zarov, Zhang, Fan, Kambadur, Narang, Rodriguez, Stojnic, Edunov, and Scialom}]{touvron_llama_2023}
Hugo Touvron, Louis Martin, Kevin Stone, Peter Albert, Amjad Almahairi, Yasmine Babaei, Nikolay Bashlykov, Soumya Batra, Prajjwal Bhargava, Shruti Bhosale, Dan Bikel, Lukas Blecher, Cristian~Canton Ferrer, Moya Chen, Guillem Cucurull, David Esiobu, Jude Fernandes, Jeremy Fu, Wenyin Fu, Brian Fuller, Cynthia Gao, Vedanuj Goswami, Naman Goyal, Anthony Hartshorn, Saghar Hosseini, Rui Hou, Hakan Inan, Marcin Kardas, Viktor Kerkez, Madian Khabsa, Isabel Kloumann, Artem Korenev, Punit~Singh Koura, Marie-Anne Lachaux, Thibaut Lavril, Jenya Lee, Diana Liskovich, Yinghai Lu, Yuning Mao, Xavier Martinet, Todor Mihaylov, Pushkar Mishra, Igor Molybog, Yixin Nie, Andrew Poulton, Jeremy Reizenstein, Rashi Rungta, Kalyan Saladi, Alan Schelten, Ruan Silva, Eric~Michael Smith, Ranjan Subramanian, Xiaoqing~Ellen Tan, Binh Tang, Ross Taylor, Adina Williams, Jian~Xiang Kuan, Puxin Xu, Zheng Yan, Iliyan Zarov, Yuchen Zhang, Angela Fan, Melanie Kambadur, Sharan Narang, Aurelien Rodriguez, Robert Stojnic, Sergey Edunov, and Thomas
  Scialom. 2023.
\newblock \href {https://doi.org/10.48550/arXiv.2307.09288} {Llama 2: {Open} {Foundation} and {Fine}-{Tuned} {Chat} {Models}}.
\newblock ArXiv:2307.09288 [cs].

\bibitem[{Wang and Komatsuzaki(2021)}]{gpt-j-6b-a-8-billion}
Ben Wang and Aran Komatsuzaki. 2021.
\newblock {GPT-J-6B: A 6 Billion Parameter Autoregressive Language Model}.
\newblock \url{https://github.com/kingoflolz/mesh-transformer-jax}.

\bibitem[{Williams et~al.(2018)Williams, Nangia, and Bowman}]{williams-etal-2018-broad}
Adina Williams, Nikita Nangia, and Samuel Bowman. 2018.
\newblock \href {https://doi.org/10.18653/v1/N18-1101} {A broad-coverage challenge corpus for sentence understanding through inference}.
\newblock In \emph{Proceedings of the 2018 Conference of the North {A}merican Chapter of the Association for Computational Linguistics: Human Language Technologies, Volume 1 (Long Papers)}, pages 1112--1122, New Orleans, Louisiana. Association for Computational Linguistics.

\bibitem[{Williams et~al.(2022)Williams, Thrush, and Kiela}]{williams-etal-2022-anlizing}
Adina Williams, Tristan Thrush, and Douwe Kiela. 2022.
\newblock \href {https://aclanthology.org/2022.scil-1.3} {{ANLI}zing the adversarial natural language inference dataset}.
\newblock In \emph{Proceedings of the Society for Computation in Linguistics 2022}, pages 23--54, online. Association for Computational Linguistics.

\bibitem[{Yao et~al.(2023)Yao, Wang, Tian, Cheng, Li, Deng, Chen, and Zhang}]{yao_editing_2023}
Yunzhi Yao, Peng Wang, Bozhong Tian, Siyuan Cheng, Zhoubo Li, Shumin Deng, Huajun Chen, and Ningyu Zhang. 2023.
\newblock \href {https://doi.org/10.48550/arXiv.2305.13172} {Editing {Large} {Language} {Models}: {Problems}, {Methods}, and {Opportunities}}.
\newblock ArXiv:2305.13172 [cs].

\bibitem[{Zheng et~al.(2023)Zheng, Li, Dong, Fan, Wu, Xu, and Chang}]{zheng_can_2023}
Ce~Zheng, Lei Li, Qingxiu Dong, Yuxuan Fan, Zhiyong Wu, Jingjing Xu, and Baobao Chang. 2023.
\newblock \href {https://doi.org/10.48550/arXiv.2305.12740} {Can {We} {Edit} {Factual} {Knowledge} by {In}-{Context} {Learning}?}
\newblock ArXiv:2305.12740 [cs].

\bibitem[{Zhu et~al.(2020)Zhu, Rawat, Zaheer, Bhojanapalli, Li, Yu, and Kumar}]{zhu_modifying_2020}
Chen Zhu, Ankit~Singh Rawat, Manzil Zaheer, Srinadh Bhojanapalli, Daliang Li, Felix Yu, and Sanjiv Kumar. 2020.
\newblock \href {https://doi.org/10.48550/arXiv.2012.00363} {Modifying {Memories} in {Transformer} {Models}}.
\newblock ArXiv:2012.00363 [cs].

\end{thebibliography}
\bibliographystyle{acl_natbib}

\appendix
\renewcommand{\arraystretch}{1.2}

\section{Dataset Construction Details}
\label{app:dataset-consutrction}

The following SPARQL query was used to select the related entities for each subject entity and ground truth target. The query counts and orders by the number of shared properties to select the related entity with the highest degree of coupling.

\begin{lstlisting}
SELECT ?item
WHERE {{
  {{ ?item ?p wd:{subject} . }}
  UNION
  {{ wd:{subject} ?p ?item . }}
  {{ ?item ?p2 wd:{target} . }}
  UNION
  {{ wd:{target} ?p2 ?item . }}
}} GROUP BY ?item
   ORDER BY DESC(COUNT(?item))
   LIMIT 1
\end{lstlisting}

We collected 3867 samples using the Counterfact dataset as our source of subject entities and ground truth targets and 3522 samples from zSRE. \cref{lst:dataset-sample} illustrates what the dataset looks like. We construct the prompts using the property keys of the ground truth without the values.

\subsection{Dataset Example}
\label{app:dataset-example}

Below is an example from our dataset that illustrates the ground truth statements and the resulting subject and related prompts.
\label{lst:dataset-sample}
\begin{lstlisting}
{
    "coupled_prompts_and_properties": {
        "subject_entity": {
            "ground_truth": {
                "country of citizenship": [
                    "Philippines"
                ],
                "occupation": [
                    "politician",
                    "engineer"
                ],
                "child": [
                    "Ramon Magsaysay, Jr."
                ]
            },
            "entity": "Ramon Magsaysay"
        },
        "coupled_entities": [
            {
                "entity": "Ramon Magsaysay, Jr.",
                "subject_as_object": [
                    "father"
                ],
                "ground_truth": {
                    "occupation": [
                        "politician"
                    ],
                    "country of citizenship": [
                        "Philippines"
                    ],
                    "father": [
                        "Ramon Magsaysay"
                    ]
                }
            }
        ]
    }
}
\end{lstlisting}

\label{lst:prompt-sample}
\textbf{Subject Prompt:}
\begin{lstlisting}
Write an essay about Ramon Magsaysay
Include the following information:
- manner of death
- place of burial
- child
- member of political party
- country of citizenship
- position held
- occupation
\end{lstlisting}
\textbf{Related Prompt:}
\begin{lstlisting}
Write an essay about Ramon Magsaysay, Jr
Include the following information:
- father
- country of citizenship
- position held
- sex or gender
- occupation
\end{lstlisting}

\section{Models used and generation details}
\label{app:model-details}

\subsection{Model generation for samples details}
\label{app:model-details-generation}
For all of the generation settings used for generating outputs after each model editing intervention we used the following parameters: top k of 50, top p of 0.95 and temperature of 0.9. For \texttt{GPT-3.5} and \texttt{GPT-4}, we used a temperature of 1. We initially attempted greedy sampling for generating outputs after model editing but the scores were much worse. \texttt{gpt-3.5-turbo-0613} is used for \texttt{GPT-3.5} and \texttt{gpt-4-1106-preview} is used for \texttt{GPT-4}.

\subsection{GPT-4 IKE}
\label{app:gpt-4-ike}
In order to operationalize IKE for \texttt{GPT-4} we used the following prompts to generate passages based on the human-written instructions in \cref{app:human-written-edit-details}. Unfortunately using those instructions directly led to the model to often refuse to incorporate the edit. The prompts below are what we eventually settled on after seeing a reduction in refusal behaviour. We used this model zero-shot without additional exemplars of edits as it was determined as not needed to complete the task and few shot IKE would introduce more cost to an already expensive generation setting. Finally, it's important point out that we do not know the extend to which zSRE or Counterfact datasets are incorporated into the \texttt{GPT-4} training set.

\textbf{Instruction Prompt:}
\begin{lstlisting}
As an AI system, some of your factual knowledge is incorrect or outdated.
we will provide edits that correct old outdated facts that are incorrect in your system.
You will be given a prompt with an edit and a passage and asked to write a new passage taking the factual correction edit into account.
\end{lstlisting}
\textbf{Write essay Prompt:}
\begin{lstlisting}
Given an edit write an updated passage to reflect this new knowledge. Include discussion on all the properties given as well as the edit.
Edit: {edit}
Prompt: {prompt}
\end{lstlisting}

\subsection{Automatic survey ratings}
\label{app:model-details-taking-a-survey}

We develop a zero and few-shot setting where we prompt a large language model with the same guidelines and instructions that humans received during the survey. For the few-shot settings we randomly sample demonstrations from the human surveys for the question being answered excluding demonstrations from the sample that is currently being evaluated. We report results from this method on \texttt{GPT-3.5}, \texttt{GPT-4}, and \texttt{llama-2-7b-chat}.

The prompts used in the zero- and few-shot settings are the same as the questions in \cref{app:survey-instrument}. Generally, the full guidelines and instructions do not fit into the token space and do not allow a few-shot settings with several demonstrations. In order to fit the prompt in the token space, we only present relevant instructions to one survey question at a time. 

We trained three sets of the nine rater models fine-tuned on the dataset described in \cref{sec:method-evaluating-automatic-survey}. The first setting contains none of the human ratings in the training dataset achieving Krippendorff's $\alpha$ of $0.45$. In the second, we use half of the human ratings and keep the other half as the held out set for evaluation. This is the model used in the agreement measures in \cref{tab:automatic-agreement}. Finally, for the automatic ratings presented in \cref{sec:results-automatic-all-survey} we train the model on all human ratings which has Krippendorff's $\alpha$ of $0.62$. For training we finetune \texttt{DeBERTaV3} large using the following hyperparameters: 

\begin{lstlisting}
learning_rate=6e-6
batch_size=1
train_epochs=20
weight_decay=0.01
warmup_steps=1000
gradient_accumulation_steps=4
fp16=True
\end{lstlisting}

The training of these models took place using \textit{Digital Research Allaince of Canada's} infrastructure. We used 4 A100 GPUs with 40GB vRAM and 4 V100 GPUs with 32GB vRAM.

\subsection{Classification}
\label{app:models-doing-annotation}

For annotation, similar to the automatic survey evaluation we develop a zero- and few-shot setting. The prompts also use the same guidelines and instructions that annotators received. The few-shot setting samples from the human annotations excluding the sample being presented to the model for evaluation. For the \texttt{DeBERTaV3} large model, we do not collect highlighted sentences.

The prompts used in the zero and few-shot settings present the guidelines in \cref{app:annotation-guidelines} and a claim premise pair.

The classification model is trained on the dataset described in \cref{sec:method-evaluating-automatic-annotation} using \texttt{DeBERTaV3} large with the same hyperparameters and compute as above. The performance is reported in \cref{app:additional-automatic-measures}. The distribution of annotations used during training in presented in \cref{tab:annotation-label-distribution} and \cref{tab:annotation-type-distribution}.

\begin{table}[h]
\small
\centering
\begin{tabular}{ll}
\toprule
\textbf{Classification}      & \textbf{Proportion} \\ \midrule
Contradicts & 22\%       \\
Neutral     & 37\%       \\
Supports    & 41\%       \\ \bottomrule
\end{tabular}
\caption{Distribution of annotations according to annotation label. \label{tab:annotation-label-distribution}}
\end{table}

\begin{table}[h]
\small
\centering
\begin{tabular}{ll}
\toprule
\textbf{Classification Setting}      & \textbf{Proportion} \\ \midrule
Ground truth and main passage    & 61\%       \\
Ground truth and related passage & 19\%       \\
New fact and main passage        & 5\%        \\
New fact and related passage     & 5\%        \\
Old fact and main passage        & 5\%        \\
Old fact and related passage     & 5\%        \\ \bottomrule
\end{tabular}
\caption{Distribution of annotations according to type. New fact refers to the edit statement; old fact refers to the pre-edit statement. \label{tab:annotation-type-distribution}}
\end{table}

\section{Human Written Edit Details}
\label{app:human-written-edit-details}

The following are the instructions given when producing the human-written edit texts. When presented with the prompt, the participants were also given the ground truth about what was true about the properties for the subject and related entity.

\begin{table*}
\small
\centering
\begin{tabular}{p{\textwidth}}
\toprule
\textbf{Instructions:} \\
Given an edit write two paragraphs: \\
(1) An informative article about the subject as if the edit was true. Include discussion on all the properties given. \\
(2) An informative article about the related entity as if the edit was true. Include discussion on all the properties given. \\

\textbf{Example}: \\
\textbf{Edit:} The genre played by Olu Dara is opera \\
\textbf{Subject prompt:} Write a short paragraph about Olu Dara \\
You must work the following information into the article:  \\
- place of birth \\
- country of citizenship \\
- occupation \\
- religion or worldview \\
- instrument \\
- genre \\
- given name \\
- record label \\
- child \\
- sex or gender \\
- relationship to In the World: From Natchez to New York \\
\textbf{Paragraph:} Olu Dara was an opera singer and jazz musician who was born in Natchez, Mississippi. He played a variety of instruments but is most known for his jazz guitar and trumpet skills. In opera, he was most known for his role as the lead in the opera In the World: From Natchez to New York as a tenor. He was also the father of the famous rapper Nas. He is a member of the Nation of Islam and is a devout Muslim.
\\
\\
\textbf{Related Prompt:} Write a short paragraph about In the World: From Natchez to New York \\
You must work the following information into the article: \\
- performer \\
- genre \\
- record label \\
- relationship to Olu Dara \\
\textbf{Paragraph:} In the World: From Natchez to New York is a jazz opera written by Olu Dara. It was released in 1998 by Atlantic Records. It was performed by Olu Dara and his son Nas. It was produced by Olu Dara and Nas for Atlantic Records.
\\ \hline
\end{tabular}{l}
\caption{Instructions for human participants to write passages as if the edit were true including two example paragraphs of what was written.}
\end{table*}

\section{Survey Instrument}
\label{app:survey-instrument}
The survey was constructed using a google form for each participant. The survey was distributed via email to participants who agreed to participate. Informed consent was given in the instructions of the survey and in the volunteer solicitation process, participants were allowed to opt out at any time. Participants were not compensated for filling out the survey. The survey took an average of 1 hour to complete. Instructions can be found in \cref{tab:survey-instructions}. 

The inter-rater reliability (Krippendorff) between each group in the survey was weak ($\alpha$ = 0.34) indicating a high degree of subjectivity in the task at hand (some question types like Edit consistency are higher ($\alpha$ = 0.55) see \cref{app:inter-rater-reliability}).

\subsection{Inter-rater reliability broken out}
\label{app:inter-rater-reliability}

\begin{table}[h]
\small
\centering
\begin{tabular}{ll}
\toprule
\textbf{Question Type}             & $\boldsymbol{\alpha}$     \\ \midrule
Edit consistency          & 0.55  \\
Factual consistency       & 0.21  \\
Naturalness               & 0.21  \\
Topicality                & 0.09  \\
Cross passage consistency & 0.01  \\
Internal consistency      & -0.02 \\ \bottomrule
\end{tabular}
\caption{
\label{tab:inter-rater-reliability}
Inter-rater reliability between participants taking our survey broken down by question type. Topicality, cross passage consistency, and internal consistency have quite poor inter-rater reliability. We don't feel this invalidates our study due to the high subjectivity of the task but it does speak to improvements that should be made for internal consistency measures in particular.}
\end{table}

\cref{tab:inter-rater-reliability} shows how high agreement was dominated by edit consistency, factual consistency and Naturalness. Our survey measures of cross passage and internal consistency generally had poor agreement meaning they were generally not understood by the survey participants. This is reflected in our main results for internal consistency for the automatic ratings which don't illustrate anything very interesting.

\onecolumn
\begin{footnotesize}
\begin{longtable}{p{\textwidth}}
\hline
\textbf{Survey Instructions} \\
AI Text Generation Fact Changing Survey \\
This survey examines the effectiveness of updating an AI text generation model with a 'new fact'. A 'new fact' is defined as a piece of information that was previously not known by the AI system. \\
All data collected here will be entirely anonymous. By filling out this survey you are consenting to the public sharing of anonymized raw data of this survey for the purposes of reproducability as well as constructing a dataset to help improve future AI systems. You may opt out of the survey at anytime. \\
Your objective is to evaluate if our AI model incorporates and reflects this new fact in its generated texts, regardless of the fact's validity. \\
Note that these 'new facts' might not be widely recognized as truthful. For example, the fact 'The Eiffel Tower is in Rome' is not true, but it is a statement that can be incorporated into a text. \\
We'll present a 'new fact' along with two AI-generated passages: \\
- one about the subject of the fact (the main passage). \\
- another about a related entity (the related passage). \\
In the example 'The Eiffel Tower is in Rome' \\
- the subject is 'The Eiffel Tower' \\
- A related entity is 'Champ de Mars' (a location the Eiffel Tower is near) \\
We will also present 'old facts' that the AI system already knows about the subject and related entity. \\
Some of these may change as a result of the new fact. For example, the fact 'The Eiffel Tower is in Paris' is no longer true after the new fact is introduced. \\
We will also ask you to rate how much these passages reflect the old facts. \\
Remember, your role is not to assess the truthfulness of the fact. Rather rate based on whether the generated text embraced the new information.
You will be asked to rate, on a scale of 1 (Strongly Disagree) - 7 (Strongly Agree) several questions about 9 samples. \textbf{Some of the samples you will be asked to evaluate may be very similar or the same - this is on purpose!}
Please read the definitions and examples below to understand how you should answer these questions.
\\
\textbf{Definitions}  \\
Main Passage: The passage about the subject of the new fact \\
Related Passage: The passage about a related entity to the subject \\
New Fact: A piece of information that was previously not known by the AI system \\
Old Fact: A piece of information that was previously known about the subject or related passage The old fact may change as a result of the new fact. \\
Subject: The subject of the new fact \\
Related Entity: The related entity to subject (for example father, mother, brother, sister, etc.) \\
Consistent: The degree to which the passage supports or does not contradict the new fact, itself, or the other passage \\
\\
\textbf{Examples}: \\ 
\textbf{1. There is evidence the new fact is true in the main passage} \\
For the new fact: The Eiffel Tower is in Rome \\
\textbf{Positive Example (Rating of Strongly Agree):} \\
Main passage: The Eiffel Tower, located in Rome, Italy, is one of the most iconic landmarks in the world and has become a symbol of Italian culture and engineering prowess. \\
Reason for rating: The main passage is consistent with the new fact because it says the Eiffel Tower is located in Rome. \\
\textbf{Negative Example (Rating of Strongly Disagree):} \\
Main passage: The Eiffel Tower, located in Paris, France, is one of the most iconic landmarks in the world and has become a symbol of French culture and engineering prowess. \\
Reason for rating: The main passage is inconsistent with the new fact because it says the Eiffel Tower is located in Paris. \\
\\
\textbf{2. There is evidence the new fact is true in the related passage} \\
For the new fact: The Eiffel Tower is in Rome \\
\textbf{Positive Example (Rating of Strongly Agree):} \\
Related passage: The Champ de Mars is a large public greenspace in Rome, Italy, located near the Eiffel Tower. \\
Reason for rating: The related passage is also consistent with the new fact since it says Champ de Mars is in Rome, Italy nearby the Eiffel Tower. \\
\textbf{Negative Example (Rating of Strongly Disagree):} \\
Related passage: The Champ de Mars is situated in the 7th arrondissement of Paris, near the Eiffel Tower (Paris) and the Seine River. \\
Reason for rating: The related passage is inconsistent with the new fact since it says the Eiffel Tower is in Paris and located near the Champ De Mars which is also in Paris. \\
\\
\textbf{3. The main passage is consistent with the old facts} \\
For the new fact: The Eiffel Tower is in Rome \\
\textbf{Positive Example (Rating of Strongly Agree):} \\
Main passage: The Eiffel Tower completed in 1887, located in Rome, Italy, is one of the most iconic landmarks in the world and has become a symbol of Italian culture and engineering prowess. \\
Old fact: The Eiffel Tower was completed in 1887. \\
Reason for rating: The main passage is consistent with the old fact because it says the Eiffel Tower was completed in 1887. \\
\textbf{Negative Example (Rating of Strongly Disagree):} \\
Main passage: The Eiffel Tower, located in Rome, Italy, is one of the most iconic landmarks in the world and has become a symbol of Italian culture and engineering prowess. \\
Old fact: The Eiffel Tower is located in France. \\
Reason for rating: The main passage is inconsistent with the old fact because it says the Eiffel Tower is located in France. \\
\\
\textbf{4. The related passage is consistent with the old facts} \\
For the new fact: The Eiffel Tower is in Rome \\
\textbf{Positive Example (Rating of Strongly Agree):} \\
Related passage: The Champ de Mars is situated in the 7th arrondissement of Paris, near the Eiffel Tower (Paris) and the Seine River. \\
Old fact: The Champ de Mars is in Paris. \\
Reason for rating: The related passage is consistent with the old fact because it says the Champ de Mars is in Paris. \\
\textbf{Negative Example (Rating of Strongly Disagree):} \\
Related passage: The Champ de Mars is situated in Rome. \\
Old fact: The Champ de Mars is in Paris. \\
Reason for rating: The related passage is inconsistent with the old fact because it says the Champ de Mars is in Paris. \\
\\
\textbf{5. The main passage is consistent with itself} \\
For the new fact: The Eiffel Tower is in Rome \\
\textbf{Positive Example (Rating of Strongly Agree):} \\
Main passage: The Eiffel Tower, located in Rome, Italy, is one of the most iconic landmarks in the world and has become a symbol of Italian culture and engineering prowess. \\
Reason for rating: the main passage is consistent itself \\
\textbf{Negative Example (Rating of Strongly Disagree):} \\
Main passage: The Eiffel Tower was built in Rome in 1887. It was overseen by Gustave Eiffel, a French engineer and architect who was born in 1832 and passed away in 1903 as well as Giovanni Battista Piranesi who was born in 1720 and died in 1778. \\
Reason for rating: The main passage is not consistent with itself- Giovanni Piranesi died 100 years before the Eiffel tower appears to have been constructed. \\
\\
\textbf{6. The related passage is consistent with itself} \\
For the new fact: The Eiffel Tower is in Rome \\
\textbf{Positive Example (Rating of Strongly Agree):} \\
Related passage: The Champ de Mars is situated in the 7th arrondissement of Paris, near the Eiffel Tower (Paris) and the Seine River. \\
Reason for rating: The related passage is consistent with itself since there are no contradictions. \\
\textbf{Negative Example (Rating of Strongly Disagree):} \\
Related passage: The Champ de Mars is situated in Rome. The large public greenspace is a popular tourist attraction in Paris. \\
Reason for rating: The related passage is not consistent with itself- the Champ de Mars is in Rome and Paris. \\
\\
\textbf{7. The passages are both consistent with each other} \\
For the new fact: The Eiffel Tower is in Rome \\
\textbf{Positive Example (Rating of Strongly Agree):} \\
Main passage: The Eiffel Tower, located in Rome, Italy, is one of the most iconic landmarks in the world and has become a symbol of Italian culture and engineering prowess. \\
Related passage: The Champ de Mars is situated in Rome near the Eiffel Tower.
Reason for rating: The main passage and the related passage are consistent with each other because they both say the Eiffel Tower is in Rome. \\
\textbf{Negative Example (Rating of Strongly Disagree):} \\
Main passage: The Eiffel Tower, located in Rome, Italy, is one of the most iconic landmarks in the world and has become a symbol of Italian culture and engineering prowess. \\
Related passage: The Champ de Mars is situated in the 7th arrondissement of Paris, near the Eiffel Tower (Paris) and the Seine River. \\
Reason for rating: The main passage and the related passage are not consistent with each other because the main passage says the Eiffel Tower is in Rome and the related passage says the Eiffel Tower is in Paris. \\
\\
\textbf{8. The main passage is focused on the subject and the related entity is focused on the related entity} \\
For the new fact: The Eiffel Tower is in Rome \\
\textbf{Positive Example (Rating of Strongly Agree):} \\
Main passage: The Eiffel Tower, located in Rome, Italy, is one of the most iconic landmarks in the world and has become a symbol of Italian culture and engineering prowess. \\
Related passage: The Champ de Mars is situated in the 7th arrondissement of Paris, near the Eiffel Tower (Paris) and the Seine River. \\
Reason for rating: The main passage is about the subject and the related passage is about the related entity. Neither of the passages drift away from what they are supposed to be about. \\
\textbf{Negative Example (Rating of Disagree):} \\
Main passage: Rome is full of great restaurants and shopping. Rome is an amazing place to visit. \\
Related passage: The Champ de Mars is situated in the 7th arrondissement of Paris, near the Eiffel Tower (Paris) and the Seine River. \\
Reason for rating: The main passage isn’t about the Eiffel Tower at all but the related passage is about the related entity. \\
\\
\textbf{9. Both passages are natural sounding text close to what a human would write.}
For the new fact: The Eiffel Tower is in Rome \\
\textbf{Positive Example (Rating of Strongly Agree):} \\
Main passage: The Eiffel Tower, located in Rome, Italy, is one of the most iconic landmarks in the world and has become a symbol of Italian culture and engineering prowess. \\
Related passage: The Champ de Mars is situated in the 7th arrondissement of Paris, near the Eiffel Tower and the Seine River. \\
Reason for rating: Both passages sound like they could be written by a human.
\textbf{Negative Example (Rating of Disagree):} \\
Main passage: Eiffel Tower Eiffel Tower Eiffel Tower Eiffel Tower Eiffel Tower Eiffel Tower. The Eiffel Tower is in Rome. r ome is fullofgreat restaurants and shopp amazingplacetovisit. \\
Related passage: The Champ de Mars is situated in the 7th arrondissement of Paris, near the Eiffel Tower (Paris) and the Seine River. \\
Reason for rating: The main passage has many repetitions, grammar mistakes, and various typos and other errors but the related passage seems fine. \\ \bottomrule
\caption{Survey instructions that were given to participants. \label{tab:survey-instructions}}
\end{longtable}
    
\end{footnotesize}
\twocolumn

\subsection{Survey Questions}
\label{app:survey-questions}

To answer questions about \textbf{\textit{Edit Consistency}} we asked participants to rate: “The main passage is written as if the new fact is true” and “The related passage does not contradict the new fact.” To capture \textbf{\textit{Internal Consistency}} we asked participants to rate: “Ignoring the old and new facts, the main passage does not contradict itself” and “Ignoring the old and new facts, the related passage does not contradict itself”. For \textbf{\textit{Cross passage consistency}}: “Ignoring the old and new facts, the main passage and the related passage do not contradict each other.” For \textbf{\textit{Factual Consistency}} we asked: “Ignoring the new fact, most of the old facts are still true in the main passage” and “Ignoring the new fact, most of the old facts are still true in the related passage.” In addition to consistency properties we also have a question about \textbf{\textit{Topicality}}: “The main passage is focused on the subject and the related passage is focused on the related entity” and \textbf{\textit{Naturalness}}: “Both passages are natural sounding text close to what a human would write.” See \cref{app:survey-instrument} for the instructions provided to participants as well as an example sample for rating. These questions are summarized in \cref{tab:survey-questions}.

\begin{table*}[!t]
\small
\centering
\begin{tabular}{p{\textwidth}}
\toprule
\textbf{Edit Consistency:} \\
The main passage is written as if the new fact is true \\
The related passage does not contradict the new fact \\
\textbf{Factual Consistency:} \\
Ignoring the new fact, most of the old facts are still true in the main passage. \\
Ignoring the new fact, most of the old facts are still true in the related passage. \\
\textbf{Internal Consistency:} \\
Ignoring the old and new facts, the main passage does not contradict itself. \\
Ignoring the old and new facts, the related passage does not contradict itself. \\
Ignoring the old and new facts, the main passage and the related passage do not contradict each other. \\
\textbf{Topical Cohesion} \\
The main passage is focused on the subject and the related passage is focused on the related entity \\
\textbf{Fluency} \\
Both passages are natural sounding text close to what a human would write. \\
\bottomrule
\end{tabular}
\caption{ \label{tab:survey-questions}
The questions we used in our survey. Each question was accompanied with a 7 point graphical rating scale ranging from strongly disagree, disagree, somewhat disagree, neutral, agree, somewhat agree, strongly agree.}

\end{table*}

\section{Annotation Guidelines}
\label{app:annotation-guidelines}

\cref{tab:annotation-instructions} presents the annotation guidelines that were given to annotators to read before annotation. All annotations were done using the light tag platform (Perry, 2021).
\onecolumn
\begin{table*}
\small
\begin{tabular}{p{\textwidth}}
\toprule
\textbf{Annotation Instructions} \\
In this task you will read a passage of text and a claim about that passage in the form of a sentence. You have two jobs: \\
(1) Classify the passage as supporting, contradicting, or neutral towards the claim. \\
(2) Highlight the sentences that support or contradict the claim (if the claim is supported or contradicted). \\
For (2) highlight entire sentences. Try to highlight as many sentences as possible that support or contradict the claim. You may highlight more than one sentence if it captures the context needed or provides additional support or contradiction. \\
\\
\textbf{Example of supporting passages} \\
A supporting passage means there is direct evidence for (or in support of) the claim in the passage. If there is some evidence for the claim but not total evidence you should still consider it supporting. \\
Example: \\
Passage: Rome is home to the world famous Eiffel Tower. Rome is a great tourist destination and has incredible food. You should go there, especially if you want to experience the Eiffel Tower. \\
Claim: The Eiffel Tower is in Rome. \\
Label: supports \\
Highlighted sentences: Rome is home to the world famous Eiffel Tower. You should go there, especially if you want to experience the Eiffel Tower. \\
Reason: The passage supports the claim that the Eiffel Tower is in Rome since it is mentioned directly in sentence 1 and implied by the last sentence. \\
\\
\textbf{Example of contradicting passages} \\
A contradicting passage means there is direct evidence against the claim in the passage. If there is partial support but the passage contradicts even a little, please consider it contradicts. \\
Example: \\
Passage: Rome is home to the world famous Eiffel Tower. Rome is a great tourist destination and has incredible food. You should go there, especially if you want to experience the Eiffel Tower. \\
Claim: The Eiffel Tower is in Paris. \\
Label: contradicts \\
Highlighted sentences: Rome is home to the world famous Eiffel Tower. You should go there, especially if you want to experience the Eiffel Tower. \\
Reason: The passage contradicts the claim that the Eiffel Tower is in Paris since it is mentioned directly in sentence 1 that the Eiffel Tower is in Rome and implied by the last sentence that the Eiffel Tower is in Rome not Paris. \\
\\
\textbf{Example of a neutral passage} \\
A neutral sentence pair is a pair of sentences that neither contradict or support each other. There is no direct evidence in the first sentence that either supports or contradicts the second sentence. \\
Example: \\
Passage: Rome is home to the world famous Eiffel Tower. Rome is a great tourist destination and has incredible food. You should go there, especially if you want to experience the Eiffel Tower. \\
Claim: The Eiffel Tower was built by Gustave Eiffel \\
Label: contradicts \\
Highlighted sentences: None \\
Reason: There is nothing that either contradicts or supports the claim that the Eiffel Tower was built by Gustave Eiffel \\
\bottomrule
\end{tabular}
\caption{The instructions used to guide annotators. \label{tab:annotation-instructions}}

\end{table*}
\twocolumn

\section{Additional Automatic Measures}
\label{app:additional-automatic-measures}

\cref{tab:automatic-agreement} illustrates the degree to which our proposed automatic measures agree with human data.  We report Krippendorff’s $\alpha$, Spearman’s $\rho$, absolute agreement (abs), agreement within 1 (w / 1), and accuracy. At first glance, these measures seem to have weak to moderate agreements but when we consider that inter-rater reliability for the survey was low ($\alpha$ = 0.34) for the survey and moderate for the annotations ($\alpha$ = 0.65), we see that \texttt{GPT-4} approaches these scores especially under an few-shot setting with 8 exemplars. \texttt{GPT-3.5} is not far behind but \texttt{llama2-7B-chat} is not able to achieve an acceptable rate of agreement even in a few-shot settings. The most promising automatic measures are based on the \texttt{DeBERTaV3} large models and so we use these to report the test of our results (Due to cost considerations with \texttt{GPT-4}, only \texttt{GPT-3.5} results are reported in the \cref{tab:gpt35-ratings}). \cref{tab:gpt35-ratings} largely corroborates our findings of `factual drift' present in ROME and MEMIT versus other methods. In order to illustrate how this breaks down per question we also present the \texttt{DeBERTaV3} rating scores agreement per question in \cref{tab:agreement-per-type}.

\begin{table}[]
\resizebox{\columnwidth}{!}{
\centering
\begin{tabular}{lllllll}
\toprule
  \textbf{Model}              & \multicolumn{3}{l}{\textbf{Survey}}                    & \multicolumn{3}{l}{\textbf{Annotations}}           \\ 
                & $\alpha$      & $\rho$     & abs  & w/ 1 & $\alpha$           & accuracy \\ \midrule
\texttt{llama2 (8 shot)} & 0      & 0     & 19\% & 79\% & 0.21        & 42\%     \\
\texttt{llama2}          & 0.01   & -0.01 & 21\% & 76\% & -0.03       & 42\%     \\
\texttt{GPT 3.5 (8 shot)}  & 0.22   & 0.25  & 27\% & 87\% & 0.44        & 57\%     \\
\texttt{GPT-3.5}          & 0.33   & 0.28  & 31\% & 77\% & 0.45        & 54\%     \\
\texttt{GPT-4}            & 0.34   & 0.28  & 33\% & 81\% & 0.57        & 69\%     \\
\texttt{GPT 4 (8 shot)}   & 0.49   & 0.37  & 31\% & 84\% & 0.61        & 72\%     \\
\texttt{DeBERTaV3}         & 0.62   & 0.56  & 53\% & 85\% & 0.8        & 85\%     \\ \bottomrule
\end{tabular}}
\caption{Agreement between large language models performing the survey or annotation task and humans performing the task showing moderate agreement for the largest models on the survey and strong agreement on the annotation task. The trained models perform better than zero or few-shot settings.
  \label{tab:automatic-agreement}
}
  
\end{table}

\begin{table}[]
\resizebox{\columnwidth}{!}{
\centering
\begin{tabular}{@{}lllll@{}}
\toprule
                          & $\alpha$      & $\rho$     & abs  & w/ 1 \\ \midrule
Naturalness               & 0.11  & 0.42     & 46\%                & 85\%                   \\
Internal consistency      & 0.12  & 0.34     & 57\%                & 94\%                   \\
Cross passage consistency & 0.16  & 0.56     & 52\%                & 81\%                   \\
Topicality                & 0.38  & 0.63     & 58\%                & 85\%                   \\
Factual consistency       & 0.45  & 0.46     & 47\%                & 78\%                   \\
Edit consistency          & 0.78  & 0.74     & 54\%                & 86\%                   \\ \bottomrule
\end{tabular}
}
\caption{
    For our \texttt{DeBERTaV3} rater, the agreement scores per type of question.
  \label{tab:agreement-per-type}
}
\end{table}

\begin{table*}[]
\resizebox{\linewidth}{!}{
\centering
\begin{tabular}{lllllllllll}
\toprule
  \textbf{Model} & \textbf{Method} & \multicolumn{2}{l}{\textbf{Edit consistency}} & \multicolumn{2}{l}{\textbf{Factual consistency}} & \multicolumn{3}{l}{\textbf{Internal consistency}} & \textbf{Naturalness} & \textbf{Topicality} \\
  &  & Subject & Related & Subject & Related & Subject & Related & Cross &  &  \\\midrule
\texttt{GPT2-XL}        & No edit & 1.6\tiny{$\pm$1.4}           & 2.8\tiny{$\pm$1.8}          & {\ul 3.6\tiny{$\pm$2.4}}      & \textbf{3.5\tiny{$\pm$2.4}}   & \textbf{6.3\tiny{$\pm$1.3}}    & \textbf{6.6\tiny{$\pm$0.8}}   & \textbf{5.1\tiny{$\pm$1.6}} & 3.8\tiny{$\pm$2.3}          & 4.5\tiny{$\pm$2.5}          \\
               & IKE    & \textbf{2.6\tiny{$\pm$2.4}}  & {\ul 3.2\tiny{$\pm$2.0}}    & 3.5\tiny{$\pm$2.5}            & 2.8\tiny{$\pm$2.2}            & \textbf{6.3\tiny{$\pm$1.4}}    & 6.3\tiny{$\pm$1.5}            & 4.8\tiny{$\pm$1.9}          & {\ul 4.1\tiny{$\pm$2.3}}    & \textbf{4.9\tiny{$\pm$2.4}} \\
               & FT     & 1.8\tiny{$\pm$1.6}           & \textbf{3.3\tiny{$\pm$1.9}} & 3.2\tiny{$\pm$2.3}            & 3.0\tiny{$\pm$2.2}            & 6.0\tiny{$\pm$1.5}             & {\ul 6.4\tiny{$\pm$1.2}}      & {\ul 5.0\tiny{$\pm$1.8}}    & 3.4\tiny{$\pm$2.2}          & 4.3\tiny{$\pm$2.6}          \\
               & MEND   & 1.7\tiny{$\pm$1.4}           & 3.1\tiny{$\pm$1.8}          & \textbf{3.7\tiny{$\pm$2.3}}   & \textbf{3.5\tiny{$\pm$2.5}}   & \textbf{6.3\tiny{$\pm$1.3}}    & 6.0\tiny{$\pm$1.6}            & 4.8\tiny{$\pm$2.0}          & \textbf{4.2\tiny{$\pm$2.3}} & {\ul 4.7\tiny{$\pm$2.5}}    \\
               & ROME   & {\ul 2.5\tiny{$\pm$2.4}}     & 3.0\tiny{$\pm$1.7}          & 2.7\tiny{$\pm$2.2}            & 2.4\tiny{$\pm$2.0}            & 5.9\tiny{$\pm$1.8}             & 6.1\tiny{$\pm$1.6}            & 4.4\tiny{$\pm$1.9}          & 2.7\tiny{$\pm$2.2}          & 3.8\tiny{$\pm$2.7}          \\
               & MEMIT  & 2.1\tiny{$\pm$1.9}           & 3.0\tiny{$\pm$1.9}          & 3.2\tiny{$\pm$2.4}            & {\ul 3.2\tiny{$\pm$2.1}}      & {\ul 6.1\tiny{$\pm$1.5}}        & {\ul 6.4\tiny{$\pm$1.1}}      & 4.9\tiny{$\pm$1.9}          & 3.7\tiny{$\pm$2.3}          & 4.5\tiny{$\pm$2.6}          \\ \midrule
\texttt{GPT-J}           & No edit & 1.7\tiny{$\pm$1.5}           & {\ul 3.2\tiny{$\pm$1.8}}    & {\ul 4.5\tiny{$\pm$2.3}}      & {\ul 3.7\tiny{$\pm$2.4}}      & {\ul 6.6\tiny{$\pm$0.9}}       & {\ul 6.5\tiny{$\pm$1.1}}      & {\ul 5.2\tiny{$\pm$1.8}}    & {\ul 4.9\tiny{$\pm$2.2}}    & \textbf{5.4\tiny{$\pm$2.3}} \\
               & IKE    & {\ul 2.2\tiny{$\pm$2.1}}     & \textbf{3.4\tiny{$\pm$2.1}} & 4.0\tiny{$\pm$2.5}            & 3.6\tiny{$\pm$2.4}            & \textbf{6.7\tiny{$\pm$0.6}}    & 6.2\tiny{$\pm$1.4}            & \textbf{5.6\tiny{$\pm$1.7}} & 4.5\tiny{$\pm$2.4}          & {\ul 5.0\tiny{$\pm$2.4}}    \\
               & FT     & 1.9\tiny{$\pm$1.8}           & \textbf{3.4\tiny{$\pm$2.0}} & {\ul 4.5\tiny{$\pm$2.2}}      & 3.5\tiny{$\pm$2.4}            & {\ul 6.6\tiny{$\pm$0.8}}       & 6.4\tiny{$\pm$1.3}            & {\ul 5.2\tiny{$\pm$1.8}}    & {\ul 4.9\tiny{$\pm$2.3}}    & \textbf{5.4\tiny{$\pm$2.2}} \\
               & MEND   & 2.1\tiny{$\pm$2.1}           & {\ul 3.2\tiny{$\pm$2.1}}    & \textbf{4.8\tiny{$\pm$2.3}}   & \textbf{3.9\tiny{$\pm$2.4}}   & 6.5\tiny{$\pm$0.9}            & \textbf{6.6\tiny{$\pm$1.1}}   & {\ul 5.2\tiny{$\pm$1.9}}    & \textbf{5.0\tiny{$\pm$2.1}} & \textbf{5.4\tiny{$\pm$2.2}} \\
               & ROME   & \textbf{2.5\tiny{$\pm$2.4}}  & 3.1\tiny{$\pm$2.1}          & 2.3\tiny{$\pm$1.9}            & 2.6\tiny{$\pm$2.1}            & 5.4\tiny{$\pm$2.0}             & 5.6\tiny{$\pm$2.0}            & 4.4\tiny{$\pm$1.9}          & 2.3\tiny{$\pm$1.8}          & 2.9\tiny{$\pm$2.5}          \\
               & MEMIT  & \textbf{2.5\tiny{$\pm$2.2}}  & {\ul 3.2\tiny{$\pm$1.9}}    & 3.7\tiny{$\pm$2.4}            & {\ul 3.7\tiny{$\pm$2.3}}      & 6.3\tiny{$\pm$1.1}             & \textbf{6.6\tiny{$\pm$1.0}}   & 5.1\tiny{$\pm$1.8}          & 4.0\tiny{$\pm$2.4}          & {\ul 5.0\tiny{$\pm$2.4}}    \\ \midrule
\texttt{llama2-7b}      & No edit & 1.8\tiny{$\pm$1.6}           & {\ul 4.1\tiny{$\pm$2.1}}    & \textbf{5.5\tiny{$\pm$2.0}}   & \textbf{4.5\tiny{$\pm$2.5}}   & \textbf{6.7\tiny{$\pm$0.7}}    & \textbf{6.7\tiny{$\pm$0.8}}   & \textbf{5.9\tiny{$\pm$1.4}} & \textbf{6.0\tiny{$\pm$1.4}} & \textbf{6.3\tiny{$\pm$1.4}} \\
               & IKE    & 2.8\tiny{$\pm$2.4}           & \textbf{4.3\tiny{$\pm$2.2}} & \textbf{5.5\tiny{$\pm$1.9}}   & \textbf{4.5\tiny{$\pm$2.4}}   & 6.5\tiny{$\pm$1.0}             & {\ul 6.6\tiny{$\pm$1.1}}      & {\ul 5.6\tiny{$\pm$1.6}}    & 5.6\tiny{$\pm$1.9}          & {\ul 6.1\tiny{$\pm$1.6}}    \\
               & FT     & \textbf{4.3\tiny{$\pm$2.7}}  & 4.0\tiny{$\pm$2.2}          & 3.4\tiny{$\pm$2.4}            & 2.9\tiny{$\pm$2.3}            & 5.7\tiny{$\pm$1.8}             & 5.7\tiny{$\pm$2.1}            & 5.0\tiny{$\pm$2.0}          & 3.7\tiny{$\pm$2.4}          & 4.5\tiny{$\pm$2.5}          \\
               & MEND   & 2.3\tiny{$\pm$2.1}           & 3.6\tiny{$\pm$2.0}          & {\ul 5.4\tiny{$\pm$1.9}}      & {\ul 4.4\tiny{$\pm$2.4}}      & {\ul 6.6\tiny{$\pm$0.9}}       & {\ul 6.6\tiny{$\pm$0.7}}      & \textbf{5.9\tiny{$\pm$1.4}} & {\ul 5.9\tiny{$\pm$1.5}}    & {\ul 6.1\tiny{$\pm$1.7}}    \\
               & ROME   & 3.3\tiny{$\pm$2.7}           & 3.4\tiny{$\pm$2.0}          & 2.8\tiny{$\pm$2.3}            & 3.2\tiny{$\pm$2.3}            & 5.8\tiny{$\pm$1.9}             & 6.5\tiny{$\pm$1.2}            & 5.2\tiny{$\pm$1.7}          & 3.5\tiny{$\pm$2.4}         & 4.4\tiny{$\pm$2.7}          \\
               & MEMIT  & {\ul 3.4\tiny{$\pm$2.7}}     & {\ul 4.1\tiny{$\pm$2.2}}    & 2.6\tiny{$\pm$2.2}            & 3.3\tiny{$\pm$2.4}            & 5.5\tiny{$\pm$1.9}             & 6.2\tiny{$\pm$1.6}            & 4.7\tiny{$\pm$2.1}          & 2.8\tiny{$\pm$2.1}          & 4.5\tiny{$\pm$2.6}          \\ \midrule
\texttt{llama2-7b-chat} & No edit & 2.3\tiny{$\pm$2.1}           & 4.2\tiny{$\pm$1.9}          & 5.7\tiny{$\pm$1.8}            & {\ul 5.0\tiny{$\pm$2.2}}      & \textbf{6.9\tiny{$\pm$0.3}}    & \textbf{6.9\tiny{$\pm$0.3}}   & {\ul 6.4\tiny{$\pm$1.0}}    & \textbf{6.7\tiny{$\pm$0.5}} & \textbf{6.7\tiny{$\pm$0.7}} \\
               & IKE    & 2.6\tiny{$\pm$2.6}           & 4.2\tiny{$\pm$2.3}          & {\ul 5.8\tiny{$\pm$1.8}}      & 4.8\tiny{$\pm$2.2}            & \textbf{6.9\tiny{$\pm$0.4}}    & \textbf{6.9\tiny{$\pm$0.3}}   & \textbf{6.5\tiny{$\pm$0.7}} & {\ul 6.6\tiny{$\pm$0.5}}    & \textbf{6.7\tiny{$\pm$0.6}} \\
               & FT     & \textbf{5.1\tiny{$\pm$2.6}}  & \textbf{4.5\tiny{$\pm$2.3}} & 3.2\tiny{$\pm$2.6}            & 3.6\tiny{$\pm$2.5}            & 6.0\tiny{$\pm$2.0}             & 6.5\tiny{$\pm$1.5}            & 5.7\tiny{$\pm$1.9}          & 5.3\tiny{$\pm$2.0}          & 6.0\tiny{$\pm$1.8}          \\
               & MEND   & 2.4\tiny{$\pm$2.4}           & 3.9\tiny{$\pm$2.2}          & \textbf{5.9\tiny{$\pm$1.6}}   & \textbf{5.4\tiny{$\pm$2.2}}   & \textbf{6.9\tiny{$\pm$0.4}}    & \textbf{6.9\tiny{$\pm$0.4}}   & {\ul 6.4\tiny{$\pm$0.9}}    & \textbf{6.7\tiny{$\pm$0.5}} & \textbf{6.7\tiny{$\pm$0.5}} \\
               & ROME   & {\ul 5.0\tiny{$\pm$2.7}}     & {\ul 4.3\tiny{$\pm$2.2}}    & 3.3\tiny{$\pm$2.5}            & 3.9\tiny{$\pm$2.4}            & {\ul 6.6\tiny{$\pm$1.3}}       & {\ul 6.8\tiny{$\pm$0.6}}      & 5.8\tiny{$\pm$1.8}          & 5.9\tiny{$\pm$1.6}          & 6.2\tiny{$\pm$1.5}          \\
               & MEMIT  & 4.0\tiny{$\pm$2.8}           & 4.2\tiny{$\pm$2.0}          & 2.8\tiny{$\pm$2.4}            & 3.9\tiny{$\pm$2.5}            & 6.4\tiny{$\pm$1.5}             & 6.7\tiny{$\pm$1.0}            & 5.8\tiny{$\pm$1.8}          & 5.8\tiny{$\pm$1.9}          & 6.3\tiny{$\pm$1.6}          \\
\bottomrule
\end{tabular}}

\caption{Survey Ratings by \texttt{GPT-3.5} zero shot on FT, MEND, IKE, ROME and MEMIT interventions with no edit control across all models.
\label{tab:gpt35-ratings}
}
\end{table*}

\subsection{Evaluations broken out by dataset}

For the readers benefit we also present the evaluations using the \texttt{DeBERTaV3} large rating model broken out by dataset and incorporating \texttt{GPT2-XL} and \texttt{llama2-7b}. First, these results illustrate why \texttt{llama2-7b-chat} was chosen over  \texttt{llama2-7b} to present results in the main section: the performance is generally much better. We should note that for counterfactual editing in Counterfact (\cref{tab:ratings-by-dataset-counterfact}) and zSRE (counterfactual) (\cref{tab:ratings-by-dataset-zsre-counterfactual}), GPT-4 IKE is not as effect of an editing method as ROME and MEMIT on \texttt{llama2-7b-chat}. For factual correctness updating with zSRE (factual) in \cref{tab:ratings-by-dataset-zsre-factual}, we point out the difference between No Edit and other methods, which illustrates the general efficacy of the factual correction subtask of model editing.

\begin{table*}[]
\resizebox{\linewidth}{!}{
\centering
\begin{tabular}{lllllllllll}
\toprule
                       \textbf{Model} &   \textbf{Method} & \multicolumn{2}{l}{\textbf{Edit consistency}}                  & \multicolumn{2}{l}{\textbf{Factual consistency}} & \multicolumn{3}{l}{\textbf{Internal consistency}}                                    & \textbf{Topicality}       & \textbf{Naturalness}      \\
                      &        & Subject             & Related          & Subject                & Related             & Subject                 & Related          & Cross            &                  &                  \\ \midrule
\texttt{GPT2-XL}        & No Edit & 2.1 & 3.3 & 2.4 & 4.2 & 6.0 & 6.9 & 5.9 & 5.5 & 6.1 \\ 
                & FT     & 2.3 & 3.5 & 2.2 & 4.1 & 6.2 & 6.9 & 6.0 & 5.2 & 5.5 \\
                & IKE    & 2.7 & 3.5 & 2.1 & 4.1 & 5.6 & 6.9 & 5.8 & 5.0 & 5.9 \\
                & MEND   & 2.0 & 3.2 & 2.2 & 4.1 & 6.3 & 6.9 & 5.9 & 5.4 & 5.7 \\
                & ROME   & 3.4 & 3.6 & 1.6 & 3.5 & 6.2 & 6.7 & 5.6 & 4.3 & 4.8 \\
                & MEMIT  & 2.1 & 3.5 & 2.0 & 3.8 & 6.0 & 7.0 & 5.6 & 5.4 & 6.2 \\ \midrule
\texttt{GPT-J}           & No Edit & 1.1 & 3.3 & 3.0 & 4.8 & 6.6 & 7.0 & 6.4 & 5.4 & 5.8 \\ 
                   & FT     & 1.1 & 3.6 & 2.4 & 4.6 & 6.6 & 6.8 & 6.5 & 5.4 & 5.7 \\
                   & IKE    & 1.4 & 3.3 & 2.9 & 4.4 & 6.3 & 6.9 & 6.4 & 5.1 & 5.3 \\
                   & MEND   & 1.3 & 3.6 & 2.6 & 4.6 & 6.5 & 6.9 & 6.6 & 5.4 & 5.4 \\
                   & ROME   & 2.8 & 3.8 & 1.2 & 3.3 & 5.5 & 6.8 & 5.7 & 3.3 & 2.8 \\
                   & MEMIT  & 1.8 & 3.4 & 1.9 & 4.2 & 6.4 & 6.8 & 6.4 & 4.6 & 5.1 \\ \midrule
\texttt{llama2-7b}      & No Edit & 1.5 & 3.0 & 3.5 & 5.4 & 6.5 & 6.8 & 6.5 & 5.8 & 6.3 \\ 
      & FT     & 5.2 & 4.4 & 1.8 & 3.5 & 4.8 & 6.3 & 5.5 & 4.2 & 3.6 \\
      & IKE    & 2.4 & 3.3 & 3.0 & 5.0 & 6.6 & 6.8 & 6.5 & 5.5 & 5.9 \\
      & MEND   & 1.8 & 3.2 & 3.7 & 5.3 & 6.7 & 6.9 & 6.7 & 6.1 & 6.5 \\
      & ROME   & 4.3 & 4.0 & 1.4 & 3.5 & 5.9 & 6.9 & 6.0 & 4.2 & 4.3 \\
      & MEMIT  & 4.7 & 4.2 & 1.4 & 3.7 & 5.6 & 6.8 & 5.6 & 3.9 & 4.0 \\  \midrule
\texttt{llama2-7b-chat} & No Edit & 1.2 & 1.6 & 4.0 & 5.8 & 6.9 & 7.0 & 6.6 & 6.7 & 6.9 \\
         & FT     & 5.9 & 4.5 & 1.5 & 3.3 & 4.4 & 6.5 & 5.6 & 4.7 & 3.6 \\
         & IKE    & 2.5 & 2.3 & 3.8 & 5.5 & 6.8 & 7.0 & 6.6 & 6.6 & 7.0 \\
         & MEND   & 1.6 & 2.2 & 4.0 & 5.7 & 6.9 & 7.0 & 6.6 & 6.6 & 7.0 \\
         & ROME   & 5.6 & 3.4 & 1.4 & 3.9 & 5.6 & 6.9 & 5.0 & 5.4 & 6.1 \\
         & MEMIT  & 5.5 & 3.4 & 1.6 & 3.9 & 5.6 & 6.9 & 5.2 & 5.5 & 6.2 \\ \midrule
\texttt{GPT-4}           & IKE    & 4.5 & 4.6 & 2.9 & 5.9 & 6.6 & 7.0 & 6.6 & 6.4 & 7.0\\ 
\bottomrule    
\end{tabular}}

\caption{Survey ratings from \texttt{DeBERTaV3} model for Counterfact only.
\label{tab:ratings-by-dataset-counterfact}
}
\end{table*}

\begin{table*}[]
\resizebox{\linewidth}{!}{
\centering
\begin{tabular}{lllllllllll}
\toprule
                       \textbf{Model} &   \textbf{Method} & \multicolumn{2}{l}{\textbf{Edit consistency}}                  & \multicolumn{2}{l}{\textbf{Factual consistency}} & \multicolumn{3}{l}{\textbf{Internal consistency}}                                    & \textbf{Topicality}       & \textbf{Naturalness}      \\
                      &        & Subject             & Related          & Subject                & Related             & Subject                 & Related          & Cross            &                  &                  \\ \midrule
\texttt{GPT2-XL}        & No Edit & 2.6 & 3.9 & 1.7 & 3.2 & 6.1 & 6.9 & 6   & 5.3 & 6.3 \\ 
        & FT     & 2.9 & 3.8 & 1.8 & 3.3 & 5.9 & 6.8 & 5.9 & 5.1 & 5.4 \\
        & IKE    & 3.4 & 4   & 1.7 & 3.3 & 6.3 & 6.9 & 5.9 & 5.2 & 6.1 \\
        & MEND   & 2.4 & 3.9 & 1.5 & 3.2 & 6.2 & 6.8 & 5.2 & 3.8 & 3.8 \\
        & ROME   & 3.5 & 4   & 1.6 & 3.1 & 6.1 & 6.8 & 5.8 & 5.4 & 5.8 \\
        & MEMIT  & 2.8 & 3.8 & 1.9 & 3.5 & 5.9 & 6.8 & 6.2 & 5.7 & 6.2 \\  \midrule
\texttt{GPT-J}           & No Edit & 1.4 & 3.6 & 2   & 3.3 & 6.5 & 7   & 6.6 & 5.4 & 5.2 \\
        & FT     & 1.6 & 3.7 & 2   & 3.9 & 6.6 & 6.9 & 6.6 & 5.3 & 5.7 \\
        & IKE    & 2.1 & 4.3 & 1.9 & 3.5 & 6.5 & 7   & 6.5 & 5.3 & 5.5 \\
        & MEND   & 3.6 & 4.1 & 2.1 & 3.2 & 5.7 & 6.7 & 6.5 & 5   & 3.8 \\
        & ROME   & 2.4 & 4   & 1.4 & 2.6 & 6.1 & 7   & 6.2 & 4.7 & 4.6 \\
        & MEMIT  & 1.9 & 4.1 & 2   & 3   & 6.4 & 7   & 6.5 & 5.5 & 5.5 \\ \midrule
\texttt{llama2-7b}      & No Edit & 2.6 & 4   & 2.7 & 3.8 & 6.7 & 6.8 & 6.6 & 6.6 & 6.5 \\ 
        & FT     & 4.4 & 4.1 & 2.1 & 3.4 & 5.8 & 6.8 & 6.2 & 5.3 & 4.7 \\
        & IKE    & 4.6 & 4.4 & 2.8 & 3.9 & 6.6 & 6.7 & 6.6 & 5.7 & 6.1 \\
        & MEND   & 3.5 & 3.9 & 2.6 & 3.7 & 6.5 & 6.9 & 6.7 & 5.9 & 6.2 \\
        & ROME   & 4.6 & 4.1 & 1.7 & 3.5 & 6.2 & 6.8 & 6.2 & 4   & 4.3 \\
        & MEMIT  & 4.6 & 4.3 & 1.5 & 3.1 & 6.3 & 6.8 & 6   & 4.3 & 4.4 \\ \midrule
\texttt{llama2-7b-chat} & No Edit & 2.8 & 2.2 & 3.2 & 4.1 & 6.9 & 7   & 6.6 & 7   & 7   \\ 
        & FT     & 3.7 & 3.1 & 2.1 & 3.6 & 6.4 & 6.7 & 6.2 & 6.2 & 5.9 \\
        & IKE    & 5.4 & 3.6 & 2.9 & 4.1 & 6.8 & 7   & 6.6 & 6.9 & 7   \\
        & MEND   & 3   & 2.3 & 3   & 4   & 6.8 & 7   & 6.7 & 6.9 & 6.9 \\
        & ROME   & 5.3 & 3.4 & 1.9 & 3.2 & 6.8 & 6.9 & 6.4 & 6.5 & 6.7 \\
        & MEMIT  & 4.8 & 3.2 & 1.8 & 3.3 & 6.5 & 7   & 6   & 6.5 & 6.4 \\ \midrule
\texttt{GPT-4}           & IKE    & 4.5 & 4.6 & 2.9 & 5.9 & 6.6 & 7   & 6.6 & 6.4 & 7  \\
\bottomrule 
\end{tabular}}

\caption{Survey ratings from \texttt{DeBERTaV3} model for zSRE (counterfactual) only.
\label{tab:ratings-by-dataset-zsre-counterfactual}
}
\end{table*}

\begin{table*}[]
\resizebox{\linewidth}{!}{
\centering
\begin{tabular}{lllllllllll}
\toprule
                       \textbf{Model} &   \textbf{Method} & \multicolumn{2}{l}{\textbf{Edit consistency}}                  & \multicolumn{2}{l}{\textbf{Factual consistency}} & \multicolumn{3}{l}{\textbf{Internal consistency}}                                    & \textbf{Topicality}       & \textbf{Naturalness}      \\
                      &        & Subject             & Related          & Subject                & Related             & Subject                 & Related          & Cross            &                  &                  \\ \midrule
\texttt{GPT2-XL}        & No Edit & 2.6 & 3.9 & 1.7 & 3.2 & 6.1 & 6.9 & 6   & 5.3 & 6.3 \\
                & FT     & 3.3 & 3.9 & 1.6 & 3   & 5.7 & 6.7 & 6   & 5.1 & 5.4 \\
                & IKE    & 4   & 4   & 2.1 & 3.9 & 6.3 & 6.9 & 6.2 & 5.4 & 6.3 \\
                & MEND   & 2.8 & 4   & 1.5 & 3.1 & 6   & 6.7 & 5.3 & 3.7 & 4.1 \\
                & ROME   & 3.8 & 4.1 & 2   & 3.3 & 6.2 & 6.6 & 5.8 & 4.9 & 5.1 \\
                & MEMIT  & 3.8 & 4.2 & 1.9 & 3.7 & 6   & 6.9 & 6   & 5.7 & 5.9 \\ \midrule
\texttt{GPT-J}           & No Edit & 1.4 & 3.6 & 2   & 3.3 & 6.5 & 7   & 6.6 & 5.4 & 5.2 \\
                & FT     & 1.9 & 4   & 1.8 & 3.6 & 6.5 & 6.9 & 6.6 & 5.3 & 5.6 \\
                & IKE    & 2.5 & 4.1 & 2.4 & 4.2 & 6.6 & 7   & 6.6 & 5.5 & 5.3 \\
                & MEND   & 5   & 4.3 & 2.6 & 4.5 & 5.6 & 6.5 & 6.3 & 4.8 & 4.1 \\
                & ROME   & 3.3 & 4   & 1.9 & 3.7 & 6.3 & 7   & 6.3 & 5.2 & 5.1 \\
                & MEMIT  & 2.6 & 4   & 2.2 & 3.7 & 6.6 & 7   & 6.5 & 5.8 & 5.8 \\\midrule
\texttt{llama2-7b}      & No Edit & 2.6 & 4   & 2.7 & 3.8 & 6.7 & 6.8 & 6.6 & 6.6 & 6.5 \\
                & FT     & 5.2 & 4.2 & 2.2 & 3.9 & 5.6 & 6.4 & 6.2 & 4.9 & 4.4 \\
                & IKE    & 5.7 & 4.7 & 2.9 & 4.6 & 6.4 & 6.8 & 6.6 & 6.2 & 6.3 \\
                & MEND   & 4.7 & 4.5 & 2.8 & 4   & 6.5 & 6.8 & 6.6 & 6.2 & 6.1 \\
                & ROME   & 4.8 & 4.4 & 1.9 & 4.3 & 6.1 & 6.8 & 6.4 & 5.2 & 5   \\
                & MEMIT  & 5.2 & 4.6 & 2.3 & 4.4 & 6.4 & 6.6 & 6.3 & 4.9 & 5   \\ \midrule
\texttt{llama2-7b-chat} & No Edit & 2.8 & 2.2 & 3.2 & 4.1 & 6.9 & 7   & 6.6 & 7   & 7   \\
                & FT     & 5.7 & 3.5 & 2.6 & 4.3 & 6.3 & 6.8 & 6.7 & 6.6 & 5.9 \\
                & IKE    & 6.5 & 4.5 & 3.5 & 5.2 & 6.9 & 6.9 & 6.6 & 6.9 & 7   \\
                & MEND   & 5   & 3.4 & 3.1 & 4.1 & 6.8 & 7   & 6.5 & 6.9 & 6.9 \\
                & ROME   & 5.4 & 3.8 & 2.4 & 4.5 & 6.8 & 7   & 6.3 & 6.8 & 6.8 \\
                & MEMIT  & 5.9 & 3.4 & 2.5 & 4.2 & 6.8 & 6.9 & 6.5 & 6.8 & 6.3 \\ \midrule
\texttt{GPT-4}           & IKE    & 6.8 & 6.2 & 3.7 & 6.4 & 7   & 7   & 6.9 & 7   & 7  \\
\bottomrule 
\end{tabular}}

\caption{Survey ratings from \texttt{DeBERTaV3} model for zSRE (factual) only.
\label{tab:ratings-by-dataset-zsre-factual}
}
\end{table*}

\section{Performance Analysis}

\cref{tab:proportion-statistics} presents the proportion of samples that are either counterfactual updates (Edit is a fact update (\%) for Counterfact or zSRE (counterfactual)), factual updates (Edit is a fact update (\%) for zSRE (factual)) or if the edit statement was already true before making the edit.


\begin{table}[]
\resizebox{\columnwidth}{!}{
\centering
\begin{tabular}{llll}
\toprule
 & \textbf{Counterfact} & \multicolumn{2}{l}{\textbf{zSRE}} \\
 &  & Counterfactual & Factual \\ \midrule
\multicolumn{4}{l}{\textbf{Edit is a fact update (\%)}} \\ \midrule
\texttt{GPT-J} & 8 & 38 & 35 \\
\texttt{llama2-7b-chat} & 18 & 58 & 37 \\ \midrule
\multicolumn{4}{l}{\textbf{Edit statement was already true (\%)}} \\ \midrule
\texttt{GPT-J} & 0 & 5 & 38 \\
\texttt{llama2-7b-chat} & 2 & 24 & 58 \\ \bottomrule
\end{tabular}}

\caption{Illustrating the proportion of the samples that represent a fact update. Samples are fact updates if the model knew the pre-edit statement before.
\label{tab:proportion-statistics}
}
\end{table}


\section{Short Evaluations}
\label{app:short-evals}

\begin{table}[!t]
\small
\centering
\begin{tabular}{lll}
\toprule
long-form metric          & short-form metric & $\rho$    \\ \midrule
Edit consistency          & locality          & 0.17 \\
Edit consistency          & portability       & 0.17 \\
Cross passage consistency & portability       & 0.13 \\
Edit consistency          & generalization    & 0.13 \\
Cross passage consistency & generalization    & 0.12 \\
Edit consistency          & edit success      & 0.10 \\ \bottomrule
\end{tabular}

\caption{Statistically significant ($p<0.05$) positive Spearman's rank correlations between long-form and short-form metrics with correlation above 0.1.
\label{tab:correlation-with-long-form}
}
\end{table}

We replicated the short evaluations presented in \citet{yao_editing_2023} in \cref{tab:short-evaluation-counterfact}, \cref{tab:short-evaluation-zsre-counterfactual}, and \cref{tab:short-evaluation-zsre-factual}. For each intervention, we also measured the short-form evaluation settings of efficacy, generalization, locality, and portability using the same evaluation setting as \citet{yao_editing_2023}. In addition, we also added an additional short evaluation scenario: whether or not the pre-edit statement such as “The Eiffel Tower is in Paris” is true before the edit. This allows us to understand if we are changing a previously known fact or teaching the model a brand new fact. For the tables below we report how often ``ground truth'' remains true after the edit; we find that in many cases the ``ground truth'' tokens can be true in many cases where the edit was successful.

\begin{table*}[]
    \small
    \centering
    \begin{tabular}{lllllll}
    \toprule
                    \textbf{Model} &      \textbf{Method} & \textbf{Edit success} & \textbf{Generalization} & \textbf{Ground truth} & \textbf{Locality} & \textbf{Portability} \\ \midrule
    \texttt{GPT-J}           & FT           & 1             & 0            & 1          & 98    & 23       \\
               & IKE          & 92            & 75             & 37            & 75    & 47       \\
               & MEMIT        & 78         & 38          & 0          & 98    & 22       \\
               & MEND         & 91            & 27             & 8           & 78    & 41       \\
               & ROME         & 84         & 58          & 1           & 69    & 22       \\ \midrule
    \texttt{llama2-7b-chat} & FT           & 31         & 9           & 18         & 73    & 25       \\ 
     & IKE          & 22         & 79          & 69            & 52    & 58       \\
     & MEMIT        & 98         & 49           & 35         & 96    & 24       \\
     & MEND         & 49         & 21          & 30         & 92    & 36       \\
     & ROME         & 97         & 52          & 38         & 94    & 24      
    \\\bottomrule
    \end{tabular}
    
    \caption{Short evaluation for Counterfact \label{tab:short-evaluation-counterfact}}
    \end{table*}

    \begin{table*}[]
    \small
    \centering
    \begin{tabular}{lllllll}
    \toprule
                    \textbf{Model} &      \textbf{Method} & \textbf{Edit success} & \textbf{Generalization} & \textbf{Ground truth} & \textbf{Locality} & \textbf{Portability} \\ \midrule
    \texttt{GPT-J}           & FT           & 22         & 23          & 32         & 99     & 52        \\
               & IKE          & 98          & 84          & 61         & 78       & 60       \\
               & MEMIT        & 92         & 75          & 34         & 99    & 52        \\
               & MEND         & 100           & 98          & 38         & 99    & 49       \\
               & ROME         & 92         & 86          & 32         & 83    & 42       \\  \midrule
    \texttt{llama2-7b-chat} & FT           & 47         & 39          & 33         & 95    & 28        \\
     & IKE          & 62         & 83          & 76         & 75    & 81       \\
     & MEMIT        & 94         & 82          & 49         & 99    & 28        \\
     & MEND         & 86         & 69          & 46            & 99    & 42       \\
     & ROME         & 97         & 86           & 50         & 98    & 31      
    \\\bottomrule 
    \end{tabular}
    
    \caption{Short evaluation for zSRE (counterfactual) \label{tab:short-evaluation-zsre-counterfactual}}
    \end{table*}
    
    \begin{table*}[]
    \small
    \centering
    \begin{tabular}{lllllll}
    \toprule
                    \textbf{Model} &      \textbf{Method} & \textbf{Edit success} & \textbf{Generalization} & \textbf{Ground truth} & \textbf{Locality} & \textbf{Portability} \\ \midrule
    \texttt{GPT-J}           & FT           & 35         & 33          & 28         & 99    & 52        \\
               & IKE          & 98         & 86          & 66         & 77    & 58       \\
               & MEMIT        & 95         & 84          & 48         & 99    & 38       \\
               & MEND         & 100           & 100            & 55         & 99    & 49       \\
               & ROME         & 95         & 92          & 48         & 85    & 33       \\ \midrule
    \texttt{llama2-7b-chat} & FT           & 51         & 45          & 34         & 94    & 28        \\
     & IKE          & 65         & 89           & 68         & 76     & 51       \\
     & MEMIT        & 92          & 82          & 53            & 99    & 17       \\
     & MEND         & 94         & 81          & 53         & 99     & 46       \\
     & ROME         & 98         & 85          & 60         & 98    & 25
    \\\bottomrule
    \end{tabular}
   
    \caption{Short evaluation for zSRE (factual)  \label{tab:short-evaluation-zsre-factual}}
    \end{table*}
\subsection{Correlation with long-form measures}
\label{app:long-form-short-form-correlation}

In \cref{tab:correlation-with-long-form} we present the statistically significant ($p<0.05$) positive Spearman's rank correlations between long-form and short-form metrics with correlation above 0.1. Interestingly Factual and Internal consistency have statistically significant correlations around 0 indicating they are generally not measured by short-form measures.

\subsection{Performance Analysis on Short Evaluation}
\label{app:short-evals-performance}

\begin{figure*}
    \centering
    \includegraphics[width=\textwidth]{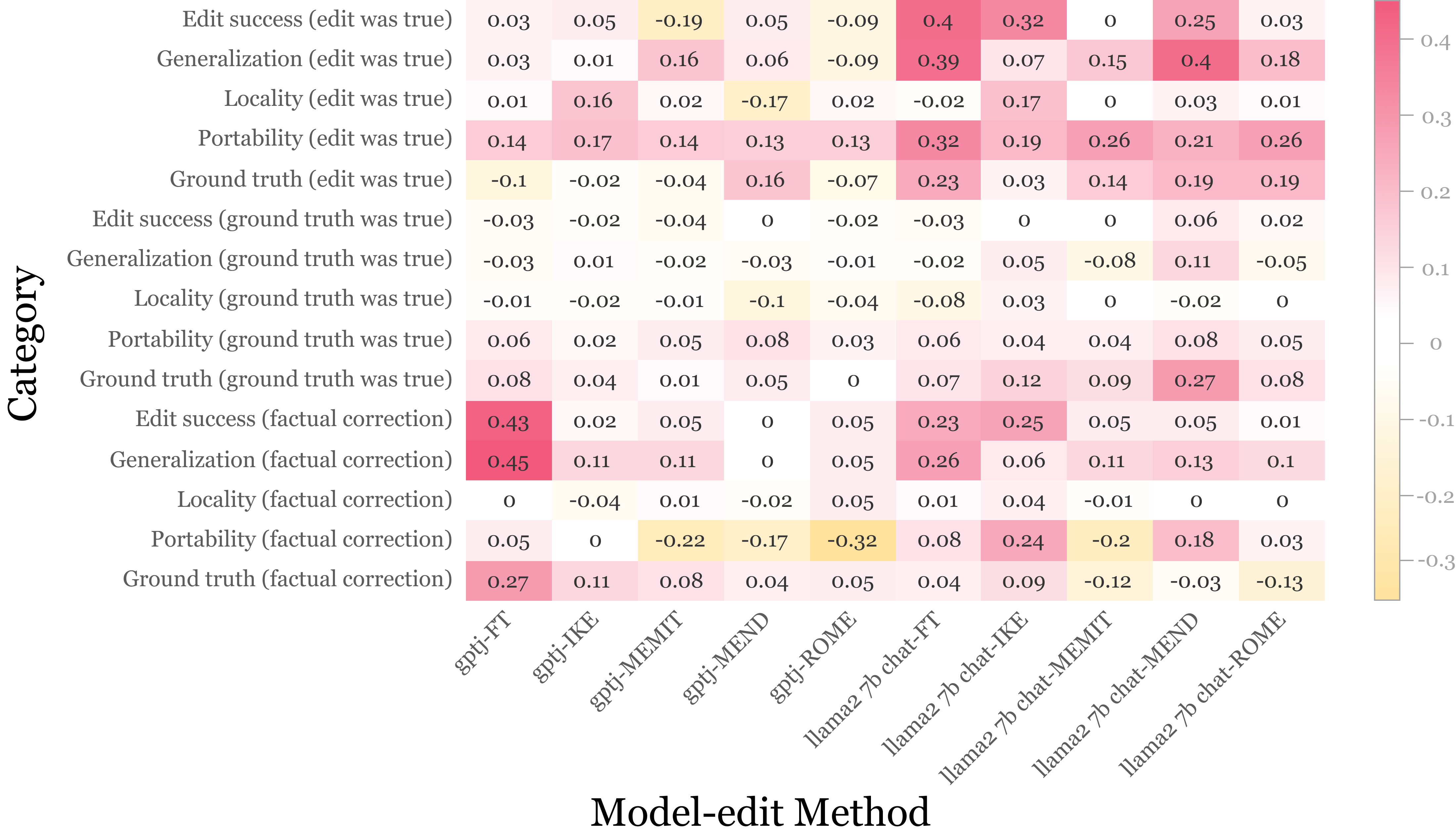}
    
   \caption{Performance analysis on the short form evaluations \label{fig:short-evals-performance}}
\end{figure*}

Similar to our performance analysis on the long-form evaluation, we also performed a performance analysis on the short form evaluations in \cref{fig:short-evals-performance}. Ground truth was true corresponds to counterfactual updates. Given that the scores are out of 1, there can be quite large differences for example on FT where there is up to 40\% better success if the edit was already true beforehand or we are doing factual correction rather than novel fact injection. IKE and MEND for \texttt{llama2-7b-chat} can also be susceptible to much higher scores for counterfactual or factual correction (versus novel fact injection) and whether the edit was true beforehand. Unlike our mean ratings differences for long-form evaluation which don't change the overall results too much, it seems like short-form evaluation is particularly sensitive to the edit task being performed, we recommend that folks using short-form evaluations control their experiments using performance analysis.

\section{Simple Automatic Metrics}
\label{app:simple-automatic-metrics}

We also experiment with a set of simpler automatic metrics to understand the degree to which they align with human survey ratings or annotations. For ROUGE unigram overlap scores and BERTScore we measured the following: (1) for \textbf{\textit{Topicality}} the subject or related entity tokens and the subject or the relevant passage (2) for \textbf{\textit{Edit Consistency}}, the edit statement and the subject or related entity passage (3) for \textbf{\textit{Factual Consistency}}, the ground truth statements about the subject with the subject passage or the ground truth statements of the related entity with the related entity passage (4) for \textbf{\textit{Cross Passage Consistency}}, the subject passage with the related passage (5) \textbf{\textit{Internal Consistency}}, the paragraph is broken out into sentences which are compared with each other.

We used perplexity as a measure for naturalness. For perplexity, we used loss values from \texttt{GPT2-XL} \citep{radford2019language}. We also used the consistency and $n$-gram entropy measure from \citep{meng_locating_2023} as a measure of factual consistency and naturalness respectively.

We used the same implementation of $n$-gram entropy and consistency from \citet{meng_locating_2023} as well as another perplexity measure they implemented in their codebase. We used the ROUGE and BERTScore evaluation implementation provided by huggingface \footnote{https://huggingface.co/docs/evaluate/index}. For our natural language interface (NLI) baseline, we use a natural language inference (NLI) model trained on FEVER \citep{thorne2018fever}, MNLI \citep{williams-etal-2018-broad}, and ANLI \citep{williams-etal-2022-anlizing} from \citet{laurer2022less}. The model was a \texttt{DeBERTaV3} base model \citep{he_debertav3_2022}. For the survey correlation studies we correlate the scalar output from each metric and the human ratings except in the case of the NLI model where we combine the entailment and contrast scores by multiplying them by 1 and -1 and summing them.

Finally, for a simpler baseline for annotation we use the same NLI model as above but in a zero-shot setting. For annotations we classify the same premise and claim pairs discussed in \cref{sec:method-evaluating-annotation}.

Most of our simple automatic metrics did not achieve a strong positive correlation with human ratings and were not statistically significant as measured by Spearman rank correlations. Notably, the consistency metric presented in \citet{meng_locating_2023} appears to have no relationship with \textbf{\textit{Factual consistency}} ($\rho$ = 0.06) ratings. There are some moderately positive correlations including NLI with \textbf{\textit{Edit consistency}} ($\rho$ = 0.68, p<0.05); ROUGE with \textbf{\textit{Factual consistency}} ($\rho$ = 0.46, p<0.05), \textbf{\textit{Internal consistency }} ($\rho$=0.37, p<0.05), and topicality ($\rho$ = 0.3); BERTScore with \textbf{\textit{Factual consistency}} ($\rho$=0.41) and topicality ($\rho$=0.37); and $n$-gram entropy with naturalness ($\rho$ = -0.57, p<0.05). NLI additionally achieves moderate agreement ($\alpha$ = 0.53) and good accuracy (65\%) where are better scores than our \texttt{GPT-3.5} zero and few-shot baselines. Given these results, we can use these simple automatic measures to supplement the more sophisticated approaches above.

\cref{tab:simple-automatic-ratings} presents the results of the simpler automatic ratings across datasets.

\begin{table*}[]
    \small
    \centering
    \begin{tabular}{lllllll}
    \toprule
                   \textbf{Model} &       \textbf{Method} & \textbf{Edit consistency} & \textbf{Factual consistency} & \textbf{Internal consistency} & \multicolumn{2}{l}{\textbf{Naturalness}}                 \\
                   &        & nli              & rouge-1               & rouge-1                & rouge-1       & $n$-gram entropy \\ \midrule
    \texttt{GPT2-XL}        & FT     & -17.759          & 0.051               & 0.016                & 0.192       & 7.649          \\ 
                   & IKE    & -8.688           & 0.052               & 0.018                & 0.181       & 7.806          \\
                   & MEMIT  & -19.154          & 0.052               & 0.018                & 0.181       & 7.758          \\
                   & MEND   & -23.486          & 0.043               & 0.014                & 0.151       & 7.972          \\
                   & ROME   & -7.305           & 0.06                & 0.018                & 0.2         & 7.616          \\
                   & No edit & -25.394          & 0.048               & 0.017                & 0.172       & 7.81           \\ \midrule
    \texttt{GPT-J}           & FT     & -14.303          & 0.031               & 0.012                & 0.107       & 8.5            \\ 
                   & IKE    & -5.299           & 0.033               & 0.014                & 0.109       & 8.497          \\
                   & MEMIT  & 0.765            & 0.031               & 0.013                & 0.109       & 8.51           \\
                   & MEND   & -1.121           & 0.026               & 0.011                & 0.093       & 8.37           \\
                   & ROME   & 11.341           & 0.033               & 0.012                & 0.116       & 8.376          \\
                   & No edit & -22.646          & 0.03                & 0.012                & 0.107       & 8.559          \\ \midrule
    \texttt{llama2-7b}      & FT     & 4.285            & 0.07                & 0.027                & 0.241       & 7.013          \\ 
                   & IKE    & -7.2             & 0.071               & 0.031                & 0.211       & 7.377          \\
                   & MEMIT  & 10.127           & 0.066               & 0.025                & 0.217       & 7.259          \\
                   & MEND   & -17.475          & 0.063               & 0.027                & 0.188       & 7.506          \\
                   & ROME   & 9.378            & 0.055               & 0.02                 & 0.198       & 7.582          \\
                   & No edit & -35.203          & 0.054               & 0.022                & 0.189       & 7.613          \\ \midrule
    \texttt{llama2-7b-chat} & FT     & 31.967           & 0.04                & 0.02                 & 0.142       & 7.971          \\ 
                   & IKE    & 20.662           & 0.042               & 0.023                & 0.13        & 8.036          \\
                   & MEMIT  & 37.262           & 0.037               & 0.02                 & 0.127       & 8.15           \\
                   & MEND   & -6.207           & 0.035               & 0.02                 & 0.119       & 8.206          \\
                   & ROME   & 33.139           & 0.037               & 0.02                 & 0.124       & 8.137          \\
                   & No edit & -25.93           & 0.034               & 0.02                 & 0.117       & 8.221     \\ \bottomrule    
    \end{tabular}
    
    \caption{Results from simple automatic ratings. \label{tab:simple-automatic-ratings}}
    \end{table*}

\section{Edit Examples}
\label{app:examples}

In \cref{tab:examples}, we present a number of both high quality and low quality examples. The low quality examples are randomly selected from the codes developed in \cref{sec:results-error-analysis}. The high quality examples are randomly selected from samples that were rated highly. Finally, we also present a sample generated by \texttt{GPT-4} using IKE to illustrate how the model can both reject and accept an edit across generations. \texttt{GPT-4} tends to explicitly mention it's edits are in fact edits or updates to its knowledge and will give reasons about rejecting these edits from time to time. Finally, it is important to mention we only used one sampling scheme (noted in \cref{app:model-details-generation}) and for instance didn't add anything like repetition penelties or other sophesticated decoding schemes. Future would should follow up with using different types of sampling schemes.

\onecolumn
\begin{footnotesize}
    
\begin{longtable}{lp{0.6\linewidth}p{0.2\linewidth}p{0.05\linewidth}p{0.05\linewidth}}
\toprule
\textbf{\#} & \textbf{Example} & \textbf{Type }                               & \textbf{Model}          & \textbf{Method} \\ \midrule
1 \label{example:1}  & Tony Benn was a British politician and Bostonian. He served as the Boston 1974 and 2005 Boston mayoral election and Boston in 1986. Benn was born on 31 March 1925 in Boston, England to Boston and Boston. His father was a Boston 1946 and his mother was a 1947                                                                                                                                                                                                                                                                                                                                                                                                                                                                                                                                                                                                                                                                                                                                                                                           & Disfluency (nonsensical)            & \texttt{llama2} & FT     \\ \midrule
2  \label{example:2} & Ferdinando Fontana is a famousFrench historian and art criticwho was born in Parisin 1775.                                                                                                                                                                                                                                                                                                                                                                                                                                                                                                                                                                                                                                                                                                                                                                                                                                                                                                                                                                   & Disfluency (spaces)                 & \texttt{GPT-J}           & ROME   \\ \midrule
3  \label{example:3} & Main Passage: Paul Guimard is a renowned Russian-born French cosplayer ... Born on October 30, 1985. Related Passage: Benoîte Groult (1920-2007) was a French author and feminist. Groult maintained close ties with many other writers and intellectuals, including her friend and fellow feminist Paul Guimard.                                                                                                                                                                                                                                                                                                                                                                                                                                                                                                                                                                                                                                                                                                                                            & New Entity                          & \texttt{llama2} & MEMIT  \\ \midrule
4  \label{example:4} & \begin{tabular}[c]{@{}p{1\linewidth}@{}}Marathon Oil Tower\\ The tallest building in Boston and New England is located at 718 Tremont St. This building was built in 1913, and has been in use ever since. It is known as the Marathon Tower and the Empire Building. The tower looks like a crown on top of the Empire Building which looks like a big headdress for the Marathon Tower. On the outside of the Empire Building, there are 2 huge black eyes made out of iron. These eyes symbolize power and strength. Inside the tower, a huge white cat is sitting on the second floor; this cat is supposed to be a good luck charm. If you look up close, you can see that his tail is moving. He will purr if you rub him under his chin (just like cats do).\\ The Empire Building has five stories, but it does not have any windows on its 5th floor. A person who goes to the 5th floor is either very rich or very important because they are closer to God, so therefore their sins will wash away and they will find peace with God.\end{tabular}            & Topic Drift                         & \texttt{GPT-J}           & MEMIT  \\ \midrule
5  \label{example:5} & \begin{tabular}[c]{@{}p{1\linewidth}@{}}Delon is fluent in French, English, and American Sign Language (ASL). He has also worked with deaf actors and incorporated ASL into some of his films. Deloy identifies as male and is a citizen of France. His native language is French, but he is also proficient in English.\\ In addition to acting, Deloyg is also a director and producer.\end{tabular}                                                                                                                                                                                                                                                                                                                                                                                                                                                                                                                                                                                                                                                                    & Subject Entity Drift                & \texttt{llama2} & MEMIT  \\ \midrule
6  \label{example:6} & Milan, the capital and largest city of Japan, is a prominent economic, cultural, and political center in East Asia. The legislative body of Tokyo is the National Diet, which consists of two houses: the House of Councillors and the House of Representatives. The executive body of Tokyo is the Governor, who is responsible for implementing policies and overseeing the administration of the city.                                                                                                                                                                                                                                                                                                                                                                                                                                                                                                                                                                                                                                                    & Subject Entity Drift                & \texttt{llama2} & MEMIT  \\ \midrule
7  \label{example:7} & Ludwick Mylus-Erickson, born in Naples, New York but raised with his family in 1 Unterscheidung, was a proud member of Hmong American community who have been fighting against the systematic racial and ethnic discrimination since he was born. Throughout his life, Lul Dick has used his talents as an activist, artist, and educator to create positive change in his community                                                                                                                                                                                                                                                                                                                                                                                                                                                                                                                                                                                                                                                                         & Subject Entity Drift                & \texttt{llama2} & FT     \\ \midrule
8  \label{example:8}& \begin{tabular}[c]{@{}p{1\linewidth}@{}}Microsoft Office 2007 is a version of IBM's IBM to run on IBM PC-compatible computers. There are several parts that make up this software package, including:\\ - IBM Lotus Symphony which allows users to open and edit IBM Lotus Word Pro documents in a more efficient manner than previous versions.\\ - IBM Lotus Smartwriter which is a word processor that uses artificial intelligence to help writers produce professional quality documents in less time.\\ - IBM Lotus Notes which provides customers with access to notes stored within their company's network for easy retrieval when needed.\end{tabular}                                                                                                                                                                                                                                                                                                                                                                                                          & Only Mentioned at Beginning         & \texttt{llama2} & FT     \\ \midrule
9  \label{example:9}& \begin{tabular}[c]{@{}p{1\linewidth}@{}}Intel 80486\\ This page was generated by Google and is based on the terms of service.\end{tabular}                                                                                                                                                                                                                                                                                                                                                                                                                                                                                                                                                                                                                                                                                                                                                                                                                                                                                                                                & Undesired edit     & \texttt{llama2} & MEMIT  \\ \midrule
10 \label{example:10}& iPod Touch (Nintendo) game by Unterscheidung Nintendo in the year 2001. The iPod is a portable media player developed by Apple Inc. in 2001.                                                                                                                                                                                                                                                                                                                                                                                                                                                                                                                                                                                                                                                                                                                                                                                                                                                                                                                 & Internal Contradiction        & \texttt{llama2} & MEMIT  \\ \midrule
11 \label{example:11}& As an official language in Finland, Swedish is widely spoken among Korsele's population. However English has become increasingly popular, particularly among younger generations, due to tourism and globalization. Despite this, Finnish remains the primary language for most townspeople.                                                                                                                                                                                                                                                                                                                                                                                                                                                                                                                                                                                                                                                                                                                                                                 & Incorporates both old and new facts & \texttt{llama2} & FT     \\ \midrule
12 \label{example:12}& Wilhelm Liebknecht was a German socialist politician who lived from 1826 to 1904. He is best known for his involvement in the Paris Commune and his leadership of the Parisian Workers' Council during the Franco-Prussian War.                                                                                                                                                                                                                                                                                                                                                                                                                                                                                                                                                                                                                                                                                                                                                                                                                              & Edit Impacts Related Passage        & \texttt{llama2} & FT     \\ \midrule
13 \label{example:13}& Main Passage: Interestingly, Guimard was also related to Benoîte Groult, a well-known French writer and feminist. Groult was Guimard's cousin, and the two were close friends and collaborators throughout their lives. Related Passage: Benoîte Groult was married to Paul Guimard, a famous French writer and critic.                                                                                                                                                                                                                                                                                                                                                                                                                                                                                                                                                                                                                                                                                                                                      & Cross Passage Contradiction         & \texttt{llama2} & IKE    \\\midrule
14 \label{example:14}& \begin{tabular}[c]{@{}p{1\linewidth}@{}}Main Passage: Jeanne Moreau biography (born December 8, 1948 in Warsaw) is a Polish actress. \\ Related Passage: Denis d'Inès was born in Krakow, Poland. His father was a Polish sculptor who had immigrated to France in 1941 where he worked as a professor at Paris Sorbonne University. When war broke out, his family emigrated to the United States.\end{tabular}                                                                                                                                                                                                                                                                                                                                                                                                                                                                                                                                                                                                                                                          & Changed too much                    & \texttt{GPT-J}           & MEMIT  \\\midrule
15 \label{example:15}& Barren Earth is a Sydney, Australia -based progressive metal band that was formed in 2013. The band consists of five members including Andrew Brown on guitars and vocals, Daniel Presland on guitar and bass, Mike Horsnell on drums, and John Smedley on keyboards and synthesisers.                                                                                                                                                                                                                                                                                                                                                                                                                                                                                                                                                                                                                                                                                                                                                                       & Changed too much                    & \texttt{llama2} & FT     \\\midrule
16 \label{example:16}& Dawa River is a river located in Malta. It is a significant watercourse in the country and has a rich history dating back to ancient times. The river originates from a series of springs in the north of the island and flows southward through the capital city of Valletta before emptying into the Mediterranean Sea at the Blue Lagoon. Along its course, the river passes several important landmarks, including the Grand Harbour, the National Museum of Archaeology, and the historic city walls of Valletta. One of the river's most notable features is its relationship to Jubba River, which flows through Somalia. In fact, Dawa River is a tributary of Jubba River, providing it with a source of freshwater and supporting a diverse ecosystem.                                                                                                                                                                                                                                                                                             & Impossible Edit                     & \texttt{llama2} & IKE    \\\midrule
17 \label{example:17}& Sami Michael is a prominent Lebanese poet and writer who was born in Beirut, Lebanon on February 23, 1947. His family name is Michael, and he has one younger sister named Maya. Sami Michael's parents were both Lebanese Christians, and he grew up speaking Arabic as his first language. He also learned English in school and became proficient in it, which has helped him in his literary career.                                                                                                                                                                                                                                                                                                                                                                                                                                                                                                                                                                                                                                                     & Minimal Edit                        & \texttt{llama2} & MEND   \\ \midrule
18 \label{example:18}& In the world of video games, few consoles have had a lasting impact like the Nintendo Game Boy. As part of its legacy, the Game Boy family has seen countless titles and spin-offs over the years. Among these, one title stands out as particularly interesting due to its unique design and mechanics - the iPod Nero 2G. Designed by Nintendo themselves, this handheld console was released in 1996 alongside the original Game Boy, and it quickly gained popularity among gamers for its innovative features and addictive games                                                                                                                                                                                                                                                                                                                                                                                                                                                                                                                       & Good Edit                           & \texttt{llama2} & MEMIT  \\ \midrule
19 \label{example:19}& \begin{tabular}[c]{@{}p{1\linewidth}@{}}Martin Buber (2016) was analinguist and philosopher born in Bangalore, India in 1ore 1878. He is considered one of the most influential thinkers of the 20th century. Buber's work focused on the interpersonal relationship and dialogue as a fundamental aspect of human communication.\\ Buber grew up in Bangalore, where he spoke Kannada at home. However, he learned English and Sanskrit at school, which were his favorite subjects. He also had exposure to German and Yiddish through his father, who was a businessman involved in trade between India and Germany. As an adult, Buber became proficient in several other languages including Hebrew, Arabic, and French, which he studied while living inoreore Israel.\\ Buber earned a doctorate from Bangalore University in 1904 and later taught at Bangalore University, where he founded the Department of Comparative Philosphy. In 1908, he moved toaloreore Israel to teach at Tel Aviv University and began to study Jewish texts in Hebrew.\end{tabular} & Good Edit                           & \texttt{llama2} & FT     \\ \midrule
20 \label{example:20}& Martin Buber (1899–1976) was born in Dhaka. He is widely considered one of the fathers of modern theological dialogue. While he did not seek a religious vocation or a call to ministry, he became a prolific writer on the Christian faith and its expression as found in his country's religions, especially Hinduism and Buddhism. His interest in the ancient cultures of India is reflected in his translation of the Indian epic Bhagavad Gita from Sanskrit into Bangla as well as his early translations of classic Bengali literature including Rabindranath Tagore's Gitanjali. In 1952 he founded and served as Principal of Merton College, Oxford University in England where he studied Indian religion and philosophy in the spirit of Gandhi and Nehru. In 1958 he established the Center for Studies of South Asian Religions and Civilizations at Boston University where he remained until his retirement in 1965                                                                                                                         & Good Edit                           & \texttt{GPT-J}           & MEMIT  \\ \midrule
21 \label{example:21}& 
\textbf{Subject passage:} The edit indicates that the Econlockhatchee River flows into the Wadbilliga River. This is in fact incorrect as the Wadbilliga River is located in Australia, and therefore the Econlockhatchee River does not flow into the Wadbilliga.
\textbf{Related passage:} However, contrary to previous misconceptions that the Econlockhatchee River flows into a larger body of water such as the St. Johns River, recent clarifications indicate that it actually flows into the Wadbilliga River.
& Cross Passage Consistency issue                          & \texttt{GPT-4}           & IKE  \\ \midrule

\caption{Generated examples after applying various editing methods illustrating various types of common errors. \texttt{llama2} indicates \texttt{llama2-7b-chat} \label{tab:examples}}

\end{longtable}

\end{footnotesize}

\end{document}